\documentclass[lettersize,journal]{IEEEtran}
\usepackage{amsmath,amsfonts,amssymb}
\usepackage{algorithm}
\usepackage{array}
\usepackage[caption=false,font=normalsize,labelfont=sf,textfont=sf]{subfig}
\usepackage{textcomp}
\usepackage{stfloats}
\usepackage{url}
\usepackage{verbatim}
\usepackage{graphicx}
\usepackage{xcolor}
\usepackage{cite}
\usepackage{algpseudocode} % algorithmicx 宏包的一部分
\usepackage{booktabs, multirow, tabularx}
\definecolor{headerbg}{RGB}{37,150,190}
\definecolor{rowhl}{RGB}{230,245,230}
\usepackage[pagebackref,breaklinks,colorlinks]{hyperref}

\newcommand{\reviseRtwo}[1]{#1}
\let\revise\reviseRone
\makeatletter
\AtBeginDocument{
  \renewcommand{\fname@algorithm}{Prompt}
}
\makeatother
\title{Semantics Disentanglement and Composition for Universal Image Coding with Efficiently LLM Reasoning and Generative Diffusion}
\author{
Jinming Liu, Yuntao Wei, Junyan Lin, Heming Sun,~\IEEEmembership{Member,~IEEE}, Shengyang Zhao, \\ Zhibo Chen, ~\IEEEmembership{Senior Member,~IEEE}, Wenjun Zeng,~\IEEEmembership{Fellow,~IEEE}, Xin Jin,~\IEEEmembership{Member,~IEEE}\\\thanks{
 Jinming Liu is with Shanghai Jiao Tong University (e-mail: jmliu206@sjtu.edu.cn). Yuntao Wei, Junyan Lin, Shengyang Zhao, Xin Jin and Wenjun Zeng are with Ningbo Key Laboratory of Spatial Intelligence and Digital Derivative, Ningbo Institute of Digital Twin, Eastern Institute of Technology, Ningbo, Zhejiang 315200, P.R. China; Zhejiang Key Laboratory of Industrial Intelligence and Digital Twin, Ningbo, Zhejiang 315200, P.R. China, Ningbo, China. Zhibo Chen is with University of Science and Technology of China, China. Heming Sun is with Yokohama National University, Japan. Corresponding Author: Xin Jin (e-mail: jinxin@eitech.edu.cn).
}
}
\begin{document}
\maketitle
\begin{abstract}
Learned image compression methods have shown impressive performance but are often highly specialized for either human perception or specific machine vision tasks. This specialization limits their versatility and requires costly retraining for new applications.  To address this, we introduce UniCodec, a universal codec built on a novel paradigm of semantic disentanglement at the encoder and compositional generation at the decoder. This framework is designed to simultaneously serve both human and machine needs, eliminating the need for task-specific retraining.
At the encoder, UniCodec leverages pre-generated, task-specific label codebooks created by a Large Language Model (LLM). For any given task, a grounding model uses the corresponding codebook to perform task-aware disentanglement, compressing only the most relevant image regions. This mechanism not only saves significant bits but is also the key to our system's rapid, zero-retraining adaptation: switching to a new task is as simple as selecting a new codebook. The decoder then performs compositional generation: it combines the compact, disentangled components with powerful priors from a generative diffusion model. This process reconstructs a high-quality, complete image optimized with rich detail for human perception and precise features for machine vision tasks. Extensive experiments demonstrate that UniCodec consistently outperforms existing methods, effectively bridging the gap between human-centric and machine-centric compression.

\end{abstract}

\begin{IEEEkeywords}
Image Compression, Image Coding for Machines, Large Language Model.
\end{IEEEkeywords}
\section{Introduction}
\begin{figure*}[t]
    \centering
    \includegraphics[width=1\linewidth]{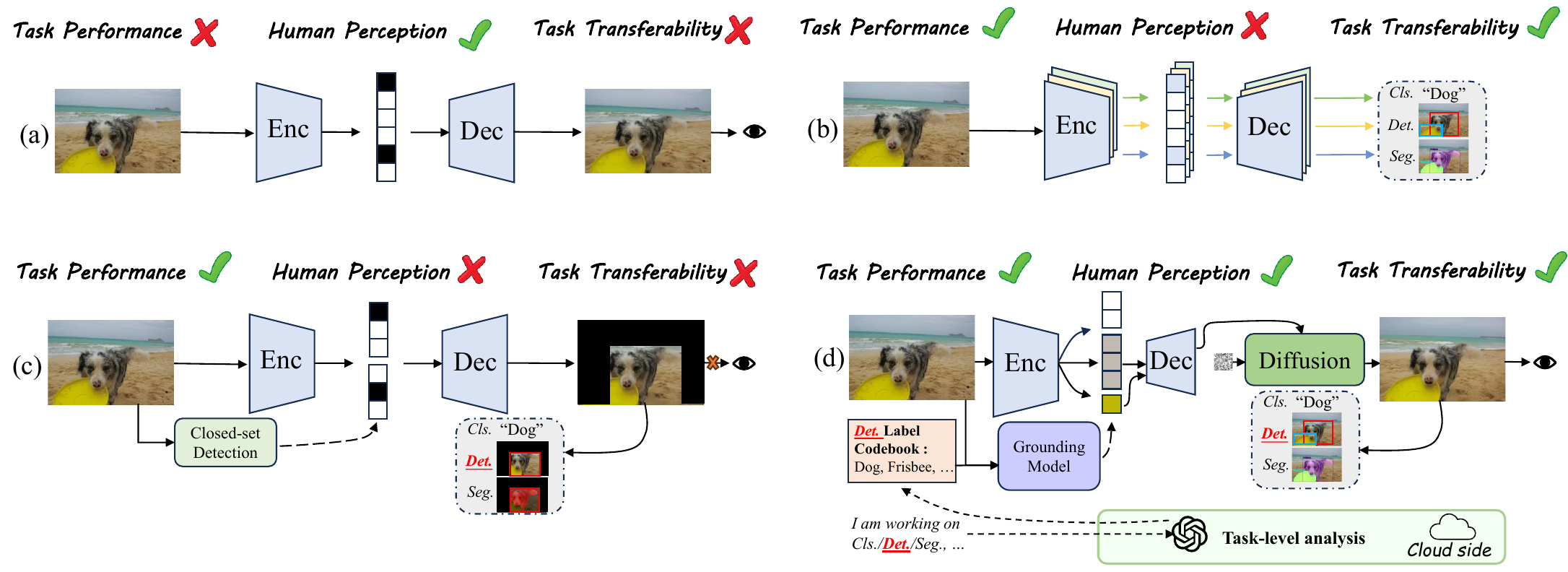}
    \caption{A comparison of image compression paradigms. (a) Traditional codecs, optimized for human perception, often struggle with machine tasks. (b, c) Conversely, machine-centric codecs improve task performance but typically sacrifice perceptual quality and transferability. (d) Our proposed UniCodec presents a universal solution. It leverages an LLM-generated codebook and a grounding model for task-aware disentanglement. At the decoder, it uses diffusion priors to compose a complete, high-quality image, successfully satisfying all three criteria: machine performance, human perception, and task transferability.}
    \label{fig:teaser}
\end{figure*}
Image compression has become an essential aspect of data management, especially as the demand for image transmission and storage continues to grow. 
Traditionally, image compression techniques like JPEG~\cite{wallace1992jpeg}, HEVC~\cite{sullivan2012overview}, VVC~\cite{bross2021overview} and advanced learned-based image compression (LIC) methods~\cite{Liu_2023_CVPR,li2023frequency,cheng2020learned} have demonstrated superior performance on standard fidelity metrics such as PSNR and MS-SSIM. For instance, Liu $et\ al.$~\cite{Liu_2023_CVPR} utilize mixed transformer-CNN architectures to capture both global and local information of the image to improve LIC, which achieves significant performance on PSNR. The latest advancements~\cite{li2024misc,agustsson2023multi,muckley2023improving,careil2023towards,li2024towards} have integrated generative models to compress images to extremely low bitrates while preserving high quality suited for human visual perception. For example, Careil et al.~\cite{careil2023towards} proposed decoding with iterative diffusion models instead of traditional feed-forward decoders~\cite{balle2017end,balle2018variational}, resulting in significantly improved visual quality. To further enhance semantic fidelity, another line of work~\cite{li2024misc,gao2024unimic} has emerged that explicitly leverages the reasoning capabilities of Multi-modal Large Language Models (MLLMs). These approaches analyze images by invoking cloud-based APIs (e.g., ChatGPT~\cite{chatgpt_2024}) or powerful local models to generate textual priors for the compression process. However, this reliance on MLLMs introduces significant practical hurdles, including substantial financial costs, latency from repeated data transmission, and high computational demands for local deployment. More fundamentally, both the diffusion-based and the MLLM-driven advancements share a critical limitation: they are exclusively optimized for human perceptual fidelity, which severely constrains their utility in the ever-expanding landscape of machine vision tasks, like object detection, segmentation, classification, etc.~\cite{ren2015faster,kirillov2023segment,radford2021learning}.

To address the needs of machine vision, which power numerous applications like autonomous driving and intelligent robotics~\cite{li2023seed,li2023one}, the field of Image Coding for Machines (ICM)~\cite{feng2022image,liu2025rate,kao2024comneck,li2024image,chen2023transtic} has emerged. ICM focuses on the joint optimization of compression rates and downstream task accuracy. Yet, these approaches represent the opposite extreme. While they excel in supporting machine tasks, they often neglect the fundamental need for a perceptually high-quality reconstruction for human viewing. Furthermore, as shown in Fig.~\ref{fig:teaser}(b), most ICM methods require costly retraining or maintaining multiple codecs to adapt to different tasks, imposing a significant overhead for practical deployment. For instance, Chen \textit{et al.}~\cite{chen2023transtic} utilizes a prompt tuning method to transfer the base codec to another machine vision task. Some alternatives, like the work by Feng et al.~\cite{feng2023prompt} shown in Fig.~\ref{fig:teaser}(c), use a pre-trained object detector to compress only foreground regions. However, its reliance on "closed-set" detectors restricts them to predefined object categories, severely hindering their transferability to new data and unseen tasks.

To resolve these dilemmas between human-centric and machine-centric coding, we introduce a fundamentally new paradigm: task-aware semantic disentanglement at the encoder, followed by generative composition at the decoder. 
% This leads to UniCodec, a universal codec designed to simultaneously satisfy both human visual perception and diverse machine vision tasks without any retraining (as shown in Fig. 1(d)). The core of UniCodec is its efficient, one-time semantic analysis: a cloud-based LLM generates a comprehensive label codebook for each task, which is then stored locally as a reusable prior. This strategy completely bypasses the prohibitive costs and latency of per-image MLLM reasoning. For established tasks, codebooks are retrieved directly from the codebook storage; for new tasks, the LLM dynamically generates new codebooks and updates the storage. This process runs only once for each task to avoid huge bandwidth and computational overhead.
Our implementation, UniCodec (as shown in Fig.~\ref{fig:teaser}(d)), embodies this principle to create a truly universal codec that adapts to diverse tasks without any retraining. 
The cornerstone of our approach is an efficient, one-time semantic analysis: a cloud-based LLM generates a comprehensive label codebook for each task, which is then stored locally as a reusable prior. This strategy—requiring only a single analysis per task, not per image—completely bypasses the prohibitive costs and latency of typical MLLM-driven methods. Furthermore, this makes our framework inherently scalable; migrating to a new task is as simple as generating a new codebook, while handling an existing one is a mere retrieval from storage.
Specifically, at the encoder, the task-specific codebook (visualized in Fig.~\ref{fig:frame}) enables precise semantic disentanglement. We feed the codebook and the input image into a visual grounding module~\cite{liu2023grounding}, which performs two critical functions: it filters out image-irrelevant labels and parses the image to extract structural representations (i.e., class identifiers and bounding boxes) of task-relevant objects. This process effectively separates the image's latent representation into two distinct parts: (1) compact, task-related semantic entities and (2) residual background information. By encoding these components separately, our architecture allows for the transmission of only the task-relevant bitstream, achieving significant bandwidth reduction and overcoming the 'closed-set' limitations of previous methods.
% Furthermore, by integrating a Codebook with a Grounding Model, we overcome the closed-set problem posed by previous methods and achieve efficient task transferability.
% The extracted label codebook (visualized in Fig.~\ref{fig:frame}) serves as input and task priors to a visual grounding module~\cite{liu2023grounding}, which performs dual functions: (1) filtering image-irrelevant task labels through image-label matching, and (2) parsing the input image into compressed task-related structural representations containing critical object-level information (including class identifiers and bounding box coordinates). Therefore, this disentanglement process can effectively separate the image into task-related semantic entities (distinct objects) and residual background components. Subsequently, we accordingly encode the disentangled objects and background separately to get a final well-structured bitstream. This architecture enables partial bitstream transmission tailored to specific downstream tasks without requiring model retraining, thereby achieving significant bandwidth reduction.
Completing the paradigm, the decoder performs semantic composition by leveraging the power of generative models~\cite{rombach2021highresolution,opensora,liu2024sora,saharia2022photorealistic}. From the transmitted partial bitstream and downsampled global information, it reconstructs the core task-related objects and coarse global information. This partial reconstruction then serves as a strong condition for a diffusion-based generative model. The diffusion model intelligently synthesizes the non-transmitted details (e.g., the background) by drawing from its vast learned priors. The final result is a visually coherent, high-quality image that not only preserves the critical information for machine interpretation but also satisfies the rich perceptual demands of human vision.
% Following this step, our decoder leverages advanced generative models~\cite{rombach2021highresolution,opensora,liu2024sora,saharia2022photorealistic} to reconstruct visually coherent full images from the transmitted partial disentangled bitstreams. The generative diffusion generates non-transmitted spatial details through learned priors and global information, ensuring both machine interpretability and human perceptual quality in the reconstructed content. 

The contributions of this paper are summarized as follows:

\begin{itemize}
    \item We introduce a new universal compression paradigm, semantic disentanglement and composition, that unifies the needs of both human perception and machine vision within a single, retraining-free framework.
    \item We design a novel encoding mechanism that leverages LLM-generated codebooks and a visual grounding module to achieve task-aware disentanglement, enabling zero-shot adaptation to new tasks and efficient partial bitstream transmission.
    \item We propose a symmetrical generative composition decoder that uses a diffusion model to synthesize high-fidelity images from compact semantic representations, ensuring quality for both human and machine interpretation. 
\end{itemize}

Experimental results demonstrate that our method achieves superior performance in both machine vision and human perception tasks. Specifically, setting VTM-12.1 as the anchor, we can achieve BD-rates~\cite{Bjntegaard2001CalculationOA} of -80.41\%, -80.32\%, and -77.63\% on object, segmentation, and classification tasks, respectively. Meanwhile, in the detection task, transmitting only the task-related bitstream achieves a BD-FID~\cite{xu2023conditional} of -45.615, indicating that the reconstructed images align well with human perception.

\section{Related Works}
\subsection{Image Compression}
Image compression aims to encode original images into a format that is both compact and retains high fidelity. Traditional image codec~\cite{wallace1992jpeg, rabbani2002overview, wiegand2003overview, sullivan2012overview, bross2021overview} has witnessed several decades of developments. In recent years, learned-based codecs~\cite{balle2017end, minnen2018joint, cheng2020learned, mentzer2018conditional, mentzer2020high, Liu_2023_CVPR,li2023frequency} began to utilize neural networks to optimize distortion and bitrate and have demonstrated superior rate-distortion performance. 
With the advancement of AI-generated content (AIGC) technologies~\cite{rombach2021highresolution,zhang2021tip,opensora}, some methods have shifted focus from image fidelity to satisfying human perception, achieving satisfactory reconstructed image quality at extremely low bitrates. 
On the other hand, as machine vision is increasingly used in practice, image coding for machine (ICM) ~\cite{he2019beyond,duan2020video,sun2020semantic,jin2023semantical,liu2022semantic,liu2023composable} is developing to satisfy machine vision tasks like image classification, detection, segmentation, etc. However, many machine task-oriented methods are typically designed for specific scenarios and often require retraining when transferred to different settings. In contrast, in this work, our proposed UniCodec can adapt to a variety of scenarios, covering both human and machine requirements, without any retraining.
% \revise{In response to the Associate Editor's request, we explicitly discuss the closest TCSVT papers here. \emph{Joint Feature and Texture Coding: Toward Smart Video Representation via Front-End Intelligence}~\cite{ma2018joint}, \emph{Semantic Structured Image Coding Framework for Multiple Intelligent Applications}~\cite{sun2020semantic}, and \emph{A Novel Video Coding Strategy in HEVC for Object Detection}~\cite{cai2021novel} all represent task-oriented semantic coding for machine analysis, but they remain tied to specific downstream semantics or detector assumptions. \emph{Towards Extreme Image Compression with Latent Feature Guidance and Diffusion Prior}~\cite{li2024towards} is the closest TCSVT generative codec on the human-perception side, yet it does not address retraining-free task adaptation for machine vision. UniCodec differs from these lines by jointly combining reusable LLM codebooks, open-vocabulary grounding, structured partial bitstreams, and diffusion-space composition so that one codec can support both human perception and multiple machine tasks without retraining.}
\subsection{Large Language Model and Visual Grounding}
Large Language Models (LLM)~\cite{touvron2023llama} and Multimodal Large Language Models (MLLM)~\cite{chatgpt_2024,liu2024visual} have experienced rapid advancements over the past few years, showcasing remarkable capabilities in understanding tasks. Some works~\cite{gao2024unimic,li2024misc} have attempted to use them to improve image compression tasks. For example, Li $et\ al.$~\cite{li2024misc} uses GPT-4 to generate descriptions of images, and utilizes them to help generate reconstructed images. However, the overhead caused by frequent API calls and running models locally makes these methods difficult to implement. In contrast, we propose an efficient method that requires only one LLM reasoning per task inference.
% TODO: Introduce large language model
Visual grounding is also close to our work, in which the task aims to locate the most relevant objects or regions in an image based on a natural language query~\cite{yang2019fast, liao2020real, yang2020improving}. The recent popular open-set Grounding DINO~\cite{liu2023grounding} has made grounding approaches more applicable to real-world scenarios, which utilize cross-modal alignment and contrastive learning methods to filter out the most relevant text during the inference process and achieve precise localization. In this paper, we leverage the open-set capabilities of Grounding DINO achieving an effective semantics analysis and disentanglement for the subsequent compression.

\subsection{Image Generation}
Recently, diffusion models~\cite{avrahami2022blended,couairon2022diffedit,meng2021sdedit,parmar2023zero} have demonstrated remarkable capabilities in image generation by the iterative denoising of a noised input. Consequently, they can generate high-quality, full images from inputs with limited information, such as text, semantic maps, or depth maps~\cite{rombach2021highresolution}. Some studies~\cite{careil2023towards,li2024misc} have attempted to apply this capability to image compression, transmitting only text and downsampled images/features to the decoder, which then employs diffusion to generate visuals that satisfy human vision. However, these approaches do not fully exploit diffusion’s potential for semantic composition and generation. In our work, we perform different semantic compositions for different tasks, to generate a full image so as to meet the needs of different scenes or tasks.

% \vspace{-2mm}
% \section{Methodology}
% \vspace{-1mm}
% \begin{figure*}
%     \centering
%     \includegraphics[width=1\linewidth]{figs/framework1.pdf}
%     \vspace{-7mm}
%     \caption{The framework of UniCodec: (1) First, we use LLM and grounding model to perform \textbf{\textit{semantic analysis}} and extract the location information of task-related objects. (2) The information is then used for \textbf{\textit{semantic disentanglement}} encoding, transmitting only task-related information. (3) Finally, we leverage the transmitted information and diffusion priors for \textbf{\textit{semantic composition generation}}.}
%     \vspace{-6mm}
%     \label{fig:frame}
% \end{figure*}

\section{Methodology}
\begin{figure*}
    \centering
    \includegraphics[width=1\linewidth]{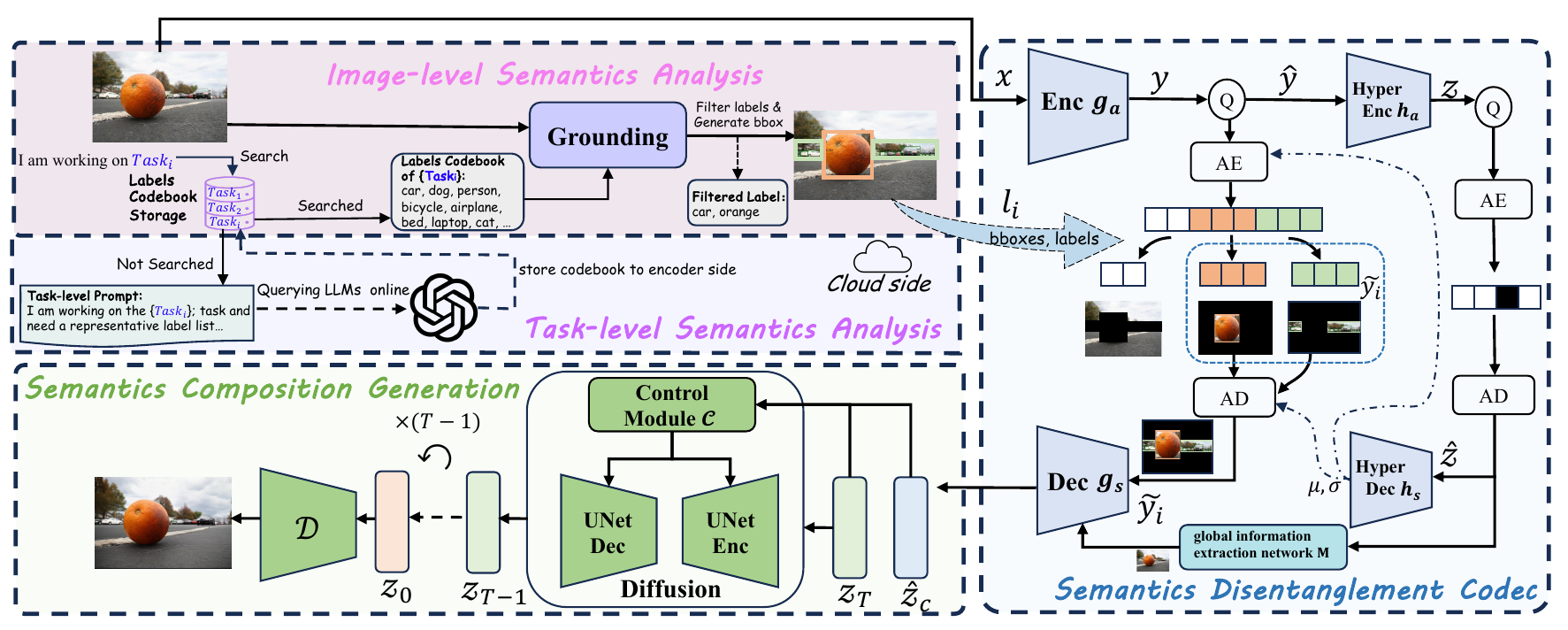}
    \caption{The framework of UniCodec: (1) \textbf{Semantic Analysis}: \reviseRtwo{utilizing an LLM to generate a label codebook for each task, which is stored at the encoder. On the encoder, for a specific \textcolor{blue}{$Task_i$}, the corresponding codebook is retrieved and used as task priors. The grounding model takes both the codebook and the original image to extract localization information of task-relevant objects.} (2) \textbf{Semantic Disentanglement Coding:} \reviseRtwo{the analysis transform $g_a$ encodes the original image $\boldsymbol{x}$ into the latent code $\hat{\boldsymbol{y}}$. The localization prior $\boldsymbol{l}_i$ from Semantic Analysis is then applied to $\hat{\boldsymbol{y}}$ via the shared operator $\mathcal{S}$ to select task-relevant latent elements $\tilde{\boldsymbol{y}}_i$.} Only task-related information and downsampled global information are compressed and transmitted. (3) \textbf{Semantic Composition Generation:} reconstructing the full image by utilizing the task-related information and diffusion priors on the decoder.}
    \label{fig:frame}
\end{figure*}

\noindent\textbf{Overview.} Different from the previous image compression methods focused exclusively on either human perception~\cite{careil2023towards,agustsson2023multi,akbari2019dsslic} or machine vision~\cite{liu2021learning,feng2022image,chen2023transtic}, our framework, as illustrated in Fig.~\ref{fig:frame}, is a universal codec for both human watching and machine tasks, and can adapt to various scenarios without retraining. 
% Firstly, it utilizes LLM and grounding modules to analyze and disentangle the semantics of the input image. This enables subsequent semantic-structured encoding, where only task-related information is encoded and transmitted for the machine tasks, significantly reducing bit overheads. Finally, the decoded image can be fully generated by composing the transmitted information with the generative diffusion prior, meeting the requirements of human watching.
The methodology unfolds in four stages. First, we detail the semantic analysis stage, where a Large Language Model and a visual grounding module work in tandem to perform task-aware image disentanglement. Second, we describe the architecture of our semantics-structured codec, which enables the partial transmission of bitstreams tailored to specific tasks. Third, we explain the generative composition process, where a diffusion-based decoder reconstructs a complete, high-fidelity image from the partial information. Finally, we introduce the specialized two-stage training strategy designed to effectively train the codec for both robust reconstruction and compositional generation.

\subsection{Semantics Analysis by LLM and Grounding}
% Previous research~\cite{liu2025rate,agustsson2023multi,korber2025egic,mentzer2020high} has demonstrated that analyzing images from a semantic perspective can better meet the requirements of human perception or machine vision at the same bit rate, compared with fidelity-based methods~\cite{Liu_2023_CVPR,li2023frequency,balle2017end,balle2018variational}.{TODO:feng,fix categ, fix task, fix format} However, these methods extract semantics solely for a single task, lacking the ability to adaptively extract diverse semantics for different scenarios, which limits their transferability across varied tasks. 

Prior works~\cite{Feng_2023_ICCV,sun2020semantic} have demonstrated that semantic priors can be used to aid downstream tasks, but their reliance on closed-set object detectors limits their transferability. To overcome this, our approach leverages the open-world understanding of large models.  We achieve this through a two-step analysis: a high-level, task-level analysis performed by an LLM, followed by a detailed, image-specific analysis using a visual grounding model~\cite{liu2023grounding}.
% In this work, we first employ an LLM to generate a label codebook for each task as a task prior. 
% Leveraging the powerful language and task-understanding capabilities of LLMs, in this work, we first employ an LLM to generate a text label codebook for each task as a task prior. Then, a grounding model~\cite{liu2023grounding} is used to perform a text-guided, one-step analysis of each image, extracting the corresponding detailed semantic information like object class and localization.

\subsubsection{Task-level Semantics Analysis by LLM}
% Large models offer exceptional versatility for a wide range of tasks. 
% Recently, some works~\cite{li2024misc,gao2024unimic} try to utilize MLLM to improve image compression tasks. However, these tasks often require significant computational resources or frequent API calls, which can lead to additional costs and bandwidth overhead. Different from those methods, as Fig.~\ref{fig:frame} shows, we guide ChatGPT-4o~\cite{chatgpt_2024} to generate a candidate semantic label codebook for each task, which provides potential task-related prior information as a bias for the subsequent process and aligns closely with task requirements. The generated label codebook can be stored in the encoder to avoid frequent calls to the LLM.
Instead of costly per-image reasoning, we employ a Large Language Model (e.g., ChatGPT-4O~\cite{chatgpt_2024}) for a one-time, task-level analysis to generate a candidate semantic label codebook, as shown in Fig.~\ref{fig:frame}. This codebook, containing potential task-related object categories, is then stored at the encoder, avoiding the need for frequent API calls or high-bandwidth data transmission. \revise{To guide the LLM, we designed a structured prompt (Prompt~\ref{alg:prompt}) that specifies a desired label count greater than 100 and the target granularity. Because this codebook is a reusable plain-text task prior with only slightly more than one hundred entries, its storage cost is negligible compared with image bitstreams.} This prompt guides the LLM through a deliberate reasoning process, first instructing it to determine the appropriate granularity for the specified task. This is refined by providing concrete examples that steer the model towards detectable object labels; for instance, encouraging general categories like 'animal' for classification to ensure robustness, while demanding specificity for other tasks. At the same time, the requirement for comprehensive coverage—over 100 labels—ensures the codebook is rich enough for diverse, open-world scenarios. The flexible nature of the prompt, with adjustable \textcolor{blue}{\textit{\{Task\}}} and \textcolor{blue}{\textit{\{Task Description\}}} fields, allows it to adapt to new applications and epitomizes the retraining-free nature of our framework.

%%%%%%%
%用公式的格式描述prompt
%%%%%%%
\begin{algorithm}
\caption{Prompt for label codebooks generation}
\begin{enumerate}
    \item Suppose you are an AI assistant and now need to generate task-related object labels based on the task.
    \item The number of labels should not be fewer than 100, and the labels should cover a wide range of categories while maintaining appropriate granularity.
    \item First, you should determine the granularity required by the task, and then generate the corresponding labels.
    \item For example:
    \begin{itemize}
        \item For some tasks, something that can't be detected, such as the sky or rivers, shouldn't be included as labels.
        \item For some tasks, it is necessary to include more comprehensive information about the image, and it needs overly general categories.
    \end{itemize}
    \item I am working on the \textcolor{blue}{\{\textit{Task}\}} and need a representative label list.
    \item The task description is as \textcolor{blue}{\{\textit{Task Description}\}}.
\end{enumerate}
\label{alg:prompt}
\end{algorithm}
% \vspace{-5mm}

% Specifically, we design a prompting method, as shown in Algorithm~\ref{alg:prompt}, which aims at guiding the model through a structured understanding of each task. The prompt firstly directs the assistant to assess the ``granularity required by the task'', which allows the assistant to generate labels at an appropriate level of granularity based on specific task requirements. After that, it requests over 100 category-specific labels, promoting both coverage and relevance. This design supports appropriate identification and categorization for the task, thereby promising both task performance and model effectiveness.
% Also, this prompt framework enables flexible task adaptation by adjusting \textcolor{blue}{\textit{\{Task\}}} and \textcolor{blue}{\textit{\{Task Description\}}}—where \textcolor{blue}{\textit{\{Task Description\}}} can be generated by ChatGPT-4o or manually defined—so we can easily switch between different tasks without reconfiguration.

\subsubsection{Image-level Semantics Analysis by Grounding}
\begin{figure}
    \centering
    \includegraphics[width=1\linewidth]{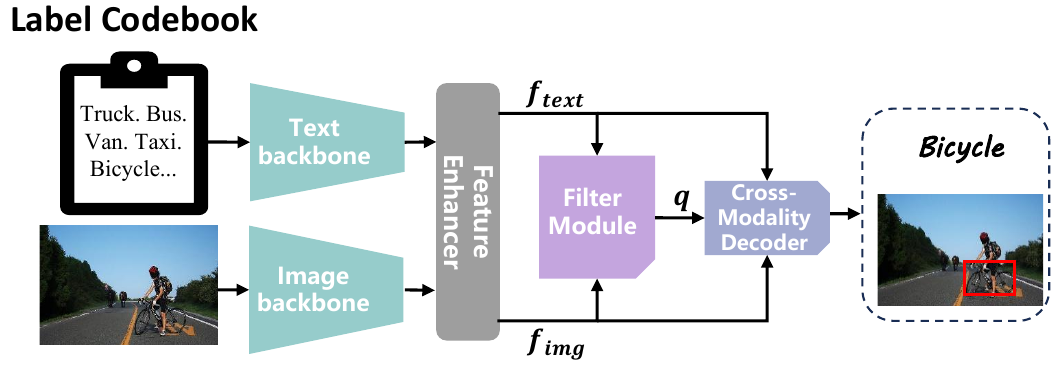}
    \caption{\revise{The label and localization generation process of the grounding module for the vehicle classification task. This design aligns the input image with the text-label codebook, enabling the extraction of codebook entries that correspond to regions present in the image. For example, in this case, it filters out the task-relevant object ``bicycle'' for subsequent compression.}}
    \label{fig:grounding}
\end{figure}

With the LLM-generated codebook serving as a rich, task-level prior, we then perform detailed image-level analysis using a visual grounding model.
Conventional closed-set object detection is restricted to a predefined vocabulary of categories, rendering it incapable of identifying novel objects and exhibiting poor scalability. In contrast, visual grounding reframes this classification task into a flexible, description-based localization. It leverages arbitrary textual descriptions to ground specific targets within an image, thereby overcoming the closed-set constraint to achieve open-vocabulary detection.
 Specifically, Grounding DINO~\cite{liu2023grounding} is utilized to filter out image-irrelevant semantic labels and retain only those most relevant to the image. Furthermore, it provides the corresponding localization of these labels within the image. \revise{In other words, the task-level codebook first provides candidate labels, and the grounding model then keeps only the labels that are relevant to the current image and outputs their corresponding boxes; these filtered labels and boxes form the localization prior used by the codec.} As shown in Fig.~\ref{fig:grounding}, we input the generated semantic labels codebook alongside the images into Grounding DINO. First, image and text features are extracted using separate backbones. These features are then processed in a feature enhancer for cross-modal fusion.

After extracting cross-modal text features $\boldsymbol{f}_{text}$ and image features $\boldsymbol{f}_{img}$, a filter module is applied that filters out superfluous and irrelevant image features under the supervision of $\boldsymbol{f}_{text}$. The selected image features are used as the cross-modality queries \( \boldsymbol{q} \), which, along with image features \( \boldsymbol{f}_{img} \) and text features \( \boldsymbol{f}_{text} \), are then input into the cross-modality decoder. Finally, the decoder outputs predicted object bounding boxes and extracts corresponding labels utilized as prior information to aid subsequent compression encoding.

\subsection{Semantics Disentanglement Codec}
After acquiring the task-related objects and regions in the image, we utilize this prior information to perform semantics disentanglement encoding, which divides the extracted objects and the background into separate groups for compression, as shown in Fig.~\ref{fig:frame}. In this way, only transmitting partial specific bitstreams, our codec can support the specific downstream machine tasks, which significantly saves bandwidth costs.

Following prior image compression works~\cite{balle2018variational,Liu_2023_CVPR}, the base codec of UniCodec comprises three components: an encoder, a hyper-prior path, and a decoder. 
Firstly, the encoder $g_a$ extracts a compact latent code \( \boldsymbol{\hat{y}} \) from the input image \( \boldsymbol{x} \).
Furthermore, using the location information $\boldsymbol{l}_i$ for task $i$ derived from the grounding model~\cite{liu2023grounding}, \( \boldsymbol{\hat{y}} \) is disentangled into different elements. When facing specific tasks, only the task-related component \( \tilde{\boldsymbol{y}}_i \) is transmitted to the decoder. On the other hand, the complete \( \boldsymbol{\hat{y}} \) is downsampled by the hyper encoder $h_a$ to create side information \( \boldsymbol{z} \), which is quantized as \( \boldsymbol{\hat{z}} \) and then transmitted to the decoder to help approximate the distribution of \( \boldsymbol{\hat{y}} \) for arithmetic coding. 
\revise{More precisely, $g_a$ always operates on the original image $\boldsymbol{x}$. The localization prior $\boldsymbol{l}_i$ is not an additional image-like input but a task-conditioned index used to project grounded boxes into latent coordinates and to select/crop/aggregate the latent elements associated with the task-relevant regions. This disentanglement operator is shared across tasks; only the grounded labels and their boxes differ from one task to another.}
The entire encoding process can be defined as:
{
\begin{equation}
\tilde{\boldsymbol{y}}_i = \mathcal{S}(g_a(\boldsymbol{x}), \boldsymbol{l}_i)\qquad \boldsymbol{\hat{z}} = h_a(g_a(\boldsymbol{x}))
\end{equation}
}
\revise{Here, $\mathcal{S}(\cdot,\boldsymbol{l}_i)$ denotes the shared latent selection/aggregation operator defined by the grounded labels and boxes for task $i$.}

On the decoder side, the coarse global information is extracted also from \( \boldsymbol{\hat{z}} \) using a global information extraction network $\mathrm{M}$ composed of two transposed convolutional layers and one convolutional layer, and is then input into the decoder $g_s$ alongside the task-related component \( \tilde{\boldsymbol{y}}_i \). It is crucial to note that our decoder's reconstruction target diverges from traditional image compression that attempts to reconstruct the original image; instead, our approach aims to reconstruct the diffusion latent code \( \boldsymbol{z}_c \) that is obtained after inputting the original image into the diffusion VAE encoder $\mathcal{E}$, thereby facilitating better integration with the diffusion model during subsequent generative composition. Ultimately, through our proposed UniCodec, we reconstruct the latent code \( \boldsymbol{\hat{z}}_c \) that encapsulates the transmitted task-related information and coarse global information. The entire decoding process can be defined as:
\begin{equation}
    \boldsymbol{\hat{z}}_c = g_s(\tilde{\boldsymbol{y}}_i, \mathrm{M}(\hat{\boldsymbol{z}}))
\end{equation}
\revise{We reconstruct the background and global layout from $\hat{\boldsymbol{z}}$ rather than directly from $\hat{\boldsymbol{y}}$ because the hyper-latent provides a lower-rate, larger-receptive-field summary of the scene, while keeping $\tilde{\boldsymbol{y}}_i$ focused on task-relevant local semantics. In our implementation, $\hat{\boldsymbol{z}}$ accounts for only about 1\%--3\% of the total bitrate, so it acts as a lightweight global condition rather than a second full-content stream.}
\revise{The target $\hat{\boldsymbol{z}}_c \approx \boldsymbol{z}_c$ also reduces the domain gap between the codec decoder and the fixed pretrained diffusion/VAE latent space. Instead of reconstructing pixels and re-encoding them, the decoder directly learns the latent domain expected by the control-conditioned diffusion model, which improves conditioning accuracy and stabilizes optimization.}

\subsection{Semantics Composition Generation}
Upon obtaining the reconstructed latent code \( \boldsymbol{\hat{z}}_c \), we employ it as a condition, which is integrated into the diffusion model's U-Net architecture~\cite{rombach2021highresolution} via a control module $\mathcal{C}$~\cite{zhang2023adding}. The control module has the same architecture as the diffusion U-Net encoder and uses several zero-convolutional layers to inject conditional knowledge into the diffusion process. The diffusion model, trained extensively on a vast dataset of images, effectively captures the distribution of images. Therefore, even with merely coarse information about the background, the diffusion model can utilize its prior knowledge to adequately inpaint this aspect, and compose the task-related objects effectively to construct a high-quality, full image, thereby satisfying human perception criteria.

Specifically, given an initial noise \( z_T \) and the condition \( \boldsymbol{\hat{z}}_c \), we employ the reverse process of the diffusion model to iteratively denoise as follows:
\begin{equation}
    \boldsymbol{z}_{t-1}=\frac{1}{\sqrt{\alpha_t}}\left(\boldsymbol{z}_t-\frac{\sqrt{1-\alpha_t}}{\sqrt{1-\bar{\alpha}_t}} \boldsymbol{\epsilon}_\theta\left(\boldsymbol{z}_t, t, \mathcal{C}(\hat{\boldsymbol{z}}_c)\right)\right)
\end{equation}
Here, \( \boldsymbol{\epsilon}_\theta(\cdot) \) represents the U-Net within the diffusion model, tasked to predict the noise at each step, with parameters denoted by \( \theta \). \( \alpha_t \) represents a scaling factor that determines how much noise is added to the latent variable \( \boldsymbol{z}_t \) at step $t$. After iterating \( T \) cycles, we obtain the denoised latent representation $\boldsymbol{z_0}$, which is then inputted into the diffusion VAE decoder $\mathcal{D}$ to produce the final reconstructed image $\boldsymbol{\hat{x}}$. 

\begin{figure}
    \centering
    \includegraphics[width=1\linewidth]{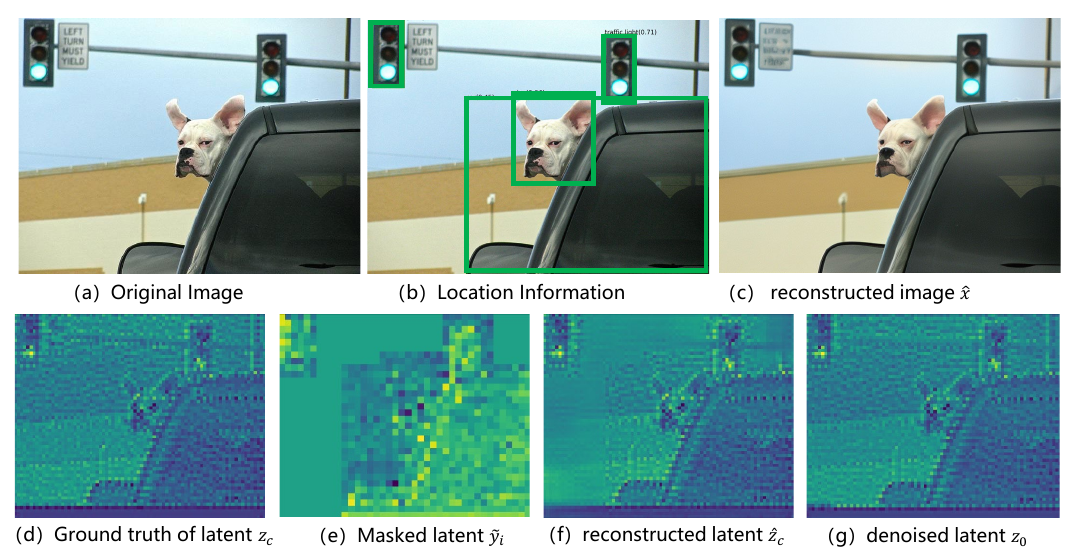}
    \caption{Visualization of the intermediate process in semantics composition generation. UniCodec can use the partial task-related compressed latent $\boldsymbol{\tilde{y}}_i$ to generate high quality image $\hat{\boldsymbol{x}}$.}
    \label{fig:latent}
\end{figure}

In Figure \ref{fig:latent}, we present several visualization results to demonstrate the feasibility of compositional generation through the task-related regions along with side priors. We attempt to reconstruct \(\boldsymbol{z}_c \), which is obtained by inputting the original image into the VAE diffusion encoder $\mathcal{E}$, as accurately as possible using the partial task-related compressed latent \( \tilde{\boldsymbol{y}_i} \). Using the information from this \( \tilde{\boldsymbol{y}}_i \) along with coarse global information from side priors \( \hat{\boldsymbol{z}} \), we can obtain a rough reconstruction of \( \hat{\boldsymbol{z}}_c \). Subsequently, leveraging the diffusion prior, we refine \( \hat{\boldsymbol{z}}_c \) to obtain a denoised latent \( \boldsymbol{z}_0 \), which is then passed through the diffusion VAE decoder to yield the reconstructed full high-quality image \( \hat{\boldsymbol{x}} \). It can be observed that the \( \boldsymbol{z}_0 \) generated by diffusion is very similar to the ground truth \( \boldsymbol{z}_c \) in Fig.~\ref{fig:latent}, thereby facilitating the production of a high-quality reconstructed image \( \hat{\boldsymbol{x}} \).

% \subsection{Joint Training of Disentanglement and Composition}
\subsection{Training Strategy}
% To implement the aforementioned pipeline for analysis, disentanglement, and composition, we divide our training steps into two stages. 
To enable the recombination of disentangled components for full image reconstruction, we design a two-stage training strategy to train a diffusion-based codec.
In the first stage, semantic analysis and disentanglement are not introduced; instead, we train a diffusion-based codec end-to-end. The specific loss function is defined as follows:
\begin{equation}
\begin{aligned}
    \mathcal{L}_I &= \lambda_R (R(\boldsymbol{\hat{y}}) + R(\boldsymbol{\hat{z}}) ) \\
    & + \lambda_{diff} \|\epsilon-\epsilon_\theta\left(\boldsymbol{z}_t, t, \mathcal{C}(\boldsymbol{\hat{z}}_c)\right\|^2 + ||\boldsymbol{z}_c - \boldsymbol{\hat{z}}_c||^2,
\end{aligned}
\label{equ:loss}
\end{equation}
where $R(\boldsymbol{\hat{y}})$ and $R(\boldsymbol{\hat{z}})$ are the rate loss, used to constrain the bitrate of the compressed bitstream. The second term measures the discrepancy between the U-Net’s predicted noise and the actual noise $\epsilon$, while the third term captures the difference between the latent code $\boldsymbol{\hat{z}}_c$ output at the decoder side and the latent code $\boldsymbol{z}_c$ obtained from the original image through the diffusion encoder $\mathcal{E}$, facilitating training stability and accelerating convergence. During this training phase, the diffusion weights, sourced from the pretrained Stable Diffusion 2.1~\cite{rombach2021highresolution}, remain frozen, and only the parameters of the control module and codec are updated.
\revise{The true noise $\epsilon$ is the Gaussian noise explicitly sampled in the forward diffusion process when constructing $\boldsymbol{z}_t$ from $\boldsymbol{z}_c$, and is therefore known during training. Stage I learns a stable latent interface between the codec and the pretrained diffusion prior under full-information inputs, while Stage II teaches the decoder to compose plausible full images from partial task-related bits plus coarse global information. We freeze the encoder and hyper-prior path in Stage II to preserve the rate-optimized latent partition and to avoid destabilizing entropy modeling.}

In the second stage of training, we attempt to endow the codec with the capabilities for disentanglement and generation. Specifically, we apply random masks to the latent code \( \boldsymbol{\hat{y}} \) during training, ensuring that only partial information along with coarse global side information is transmitted to the decoder, allowing the model to generate based on this partial information. To maintain training stability in this phase, the parameters of the encoder and the hyper-prior path within the codec remain frozen, and only the decoder is updated. The diffusion component still undergoes updates solely in the control module. The difference in the loss function from the first stage lies in the replacement of \( R(\boldsymbol{\hat{y}}) \) with rate loss for partial region parts \( R(\tilde{\boldsymbol{y}}) \).
\reviseRtwo{It is important to note that the random masking in Stage~II serves as a \textbf{task-agnostic training augmentation}: by exposing the decoder to arbitrary partial inputs, it learns a general-purpose compositional generation capability that is not tied to any specific spatial pattern. At inference time, the actual mask is determined by the LLM-Grounding module, which provides \textbf{task-aware semantic selection} of the regions that matter for the target downstream task. Section~\ref{sec:random_mask_discuss} provides a detailed discussion and ablation study comparing these two masking strategies.}
\begin{figure*}[t]
    \centering
    \includegraphics[width=1\linewidth]{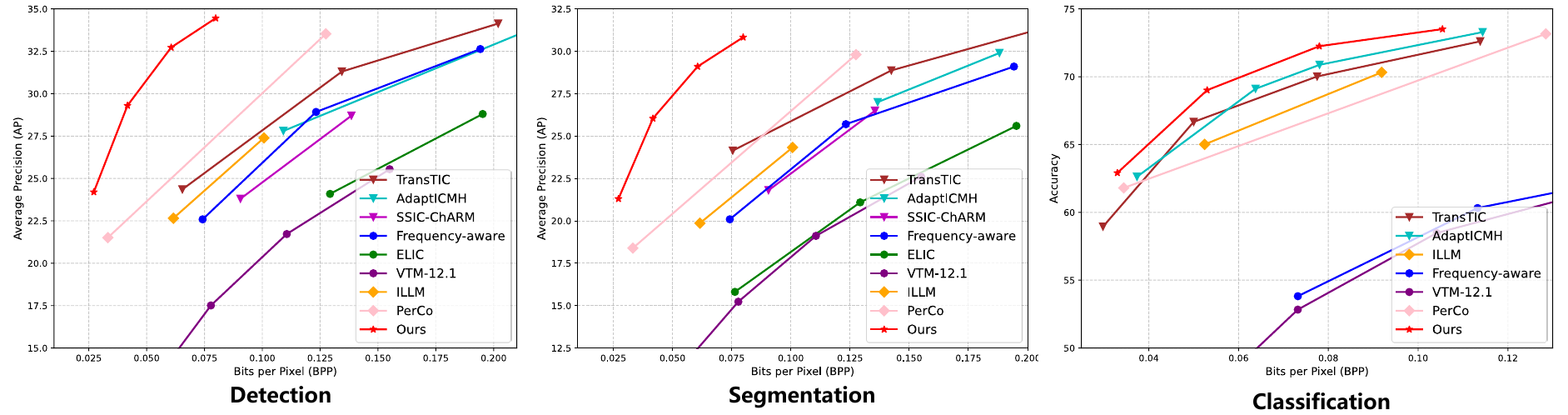}
    \caption{Machine vision tasks performance comparison. We use $\diamond$, $\triangledown$, $\circ$ to represent generative, image coding for machine, fidelity-based methods, respectively. Our method is represented by $\star$, and the proposed versatile method outperforms recent methods across three machine vision tasks without retraining while maintaining satisfactory human perception, as shown in Fig.~\ref{fig:human}.}
    \label{fig:machine}
\end{figure*}

\section{Experiments}
\label{sec:exp}
\subsection{Experimental Settings}
\subsubsection{Implementation Details}
We utilize ChatGPT-4o~\cite{chatgpt_2024} and GroundingDINO 1.0~\cite{liu2023grounding} (Swin-Tiny backbone) as the models for semantic analysis, keeping them frozen throughout the process without involving them in training stages. ELIC~\cite{he2022elic} serves as our base codec. The last two layers of its decoder are removed to align with the size of the diffusion latent code. In both training stages I and II, \(\lambda_{\text{diff}}\) is set to 2, while \(\lambda_{R}\) is chosen from \{1, 2, 4, 8\} in Equation \ref{equ:loss} to achieve models with varying bitrates. For the diffusion model, we employ Stable Diffusion 2.1~\cite{rombach2021highresolution} as the generative model, and the sampling step is set as 8 for machine vision tasks, 50 for human perception tasks. In Stage II, to enable the codec with disentangled and compositional generation capabilities, we randomly set 0$\sim$50\% of the region in \( \hat{\boldsymbol{y}} \) to zero during each forward pass before transmission to decoder. The training dataset comprises 50K images from OpenImage~\cite{OpenImages}, alongside almost 10K high-resolution images from COCO~\cite{lin2014microsoft}, CLIC training set \cite{clic}, Flickr2K~\cite{Lim_2017_CVPR_Workshops}, and Div2K~\cite{Agustsson_2017_CVPR_Workshops}. During the training, these images are randomly cropped to a size of 512x512. In stage I, the model is trained for 300,000 steps, followed by an additional 50,000 steps in Stage II. 
\revise{Unless otherwise stated, all downstream networks are evaluated with their public pretrained weights and are not retrained or finetuned on reconstructed images. This setting is intentionally chosen to test whether UniCodec preserves transferable semantics under a fixed downstream model.}
% \subsubsection{Datasets}
% \subsubsection{Machine Vision}
\begin{figure*}[!ht]
    \centering
    \includegraphics[width=1\linewidth]{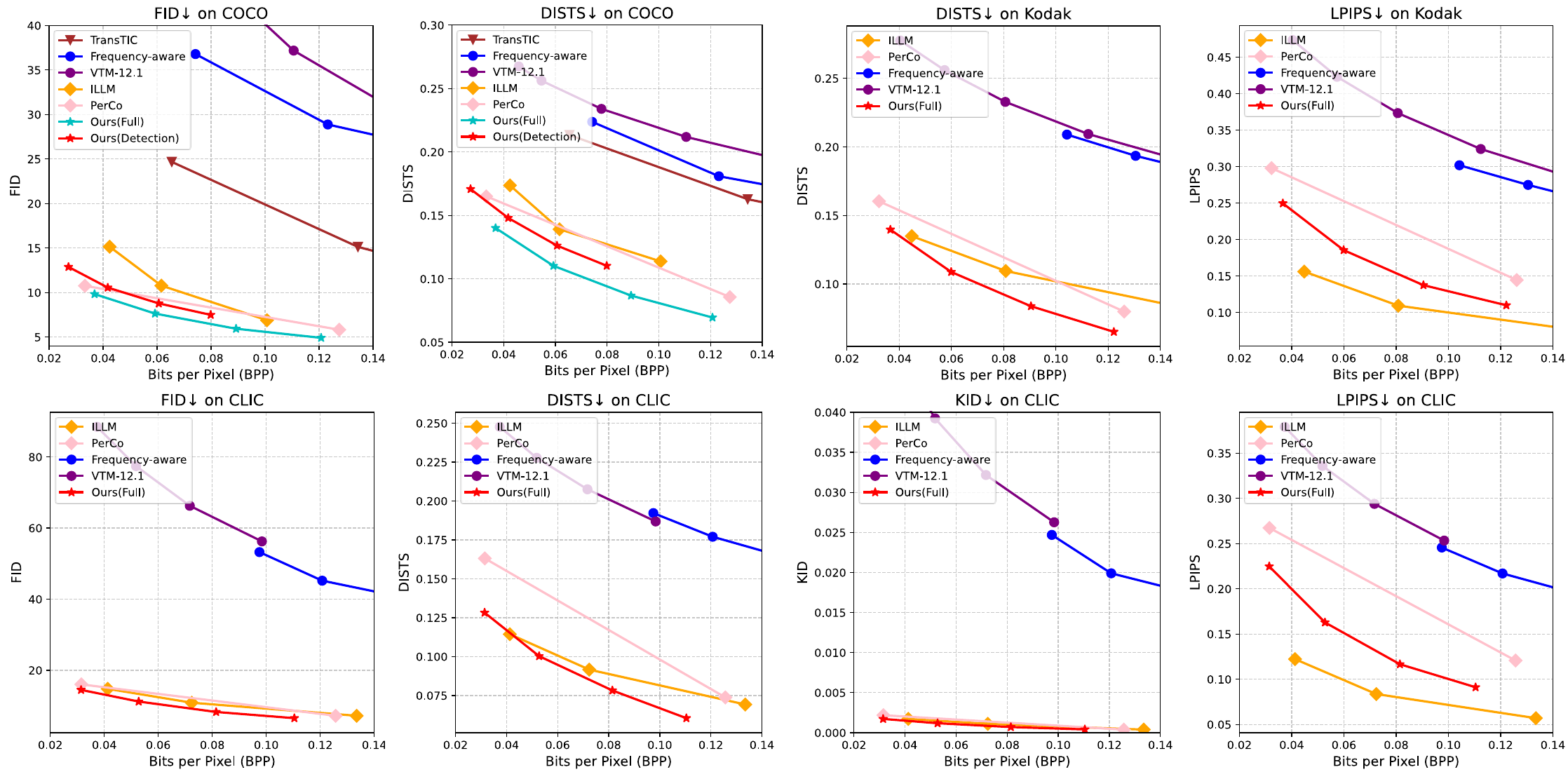}
    \caption{Human perception comparison. ``Ours(Full)" means that all bitstreams are transmitted for reconstruction, while ``Ours(Detection)" means that only disentangled task-related parts for detection task are transmitted.}
    \label{fig:human}
\end{figure*}

\paragraph{Machine Vision.} For machine vision tasks, we conduct evaluations on object detection, instance segmentation, and classification tasks. For detection and segmentation tasks, we use the MS COCO2017 val dataset with Faster R-CNN~\cite{ren2015faster} and Mask RCNN~\cite{he2017mask} (R50-FPN) as downstream task networks. For the classification task, we test ResNet-50~\cite{he2016deep} on ImageNet~\cite{deng2009large}. We also assessed the transferability of our method through brief experiments on fine-grained aircraft classification on FGVC-aircraft dataset~\cite{maji13fine-grained} and image captioning on COCO2014 dataset.

\paragraph{Human Perception.} For human perception, we performed evaluations on the Kodak \cite{kodak}, CLIC2020 \cite{clic}, and COCO2017 val dataset. Following the setting in~\cite{yang2024lossy}, for CLIC2020, images are resized so that the shorter edge is 768 pixels, followed by a center crop of 768x768.
\subsubsection{Evaluation Protocol}
\paragraph{Compression Ratio.} In this work, bits per pixel (bpp) is used to measure the compression ratio. Following~\cite{liu2025rate}, bpp is formulated as: $\frac{b}{p}$, where $b$ denotes the total bits cost of all encoded bit-streams in the datasets. $p$ represents the total pixels of the images in the dataset. 

\paragraph{Machine Vision.} In terms of evaluation metrics, for machine vision, we used Average Precision (AP) to evaluate performance in object detection and instance segmentation tasks, and accuracy for classification tasks. 

\paragraph{Human Perception.} For human perception, we used FID~\cite{heusel2017gans}, KID~\cite{binkowski2018demystifying}, DISTS~\cite{ding2020image}, and LPIPS~\cite{zhang2018unreasonable} as evaluation metrics. Similar to~\cite{mentzer2020high}, the images from CLIC are divided into 256x256 patches to compute FID and KID.

\subsubsection{Comparison Approaches}
We conducted comparisons with three categories of methods: image coding for machine methods (TransTIC~\cite{chen2023transtic}, Adapt-ICMH~\cite{li2024image}, SSIC-ChARM~\cite{Feng_2023_ICCV}), which are designed for machine tasks, generative image compression methods (ILLM~\cite{muckley2023improving}, PerCo~\cite{careil2023towards}) tailored for human perception, and some fidelity-based approaches (Frequency-aware~\cite{li2023frequency}, ELIC~\cite{he2022elic} and VTM-12.1~\cite{vvc}).

\subsection{Performance Comparison}
\subsubsection{Machine Vision}
As illustrated in Fig.~\ref{fig:machine}, we performed the comparison of machine vision tasks, including detection, segmentation, and classification. The SOTA diffusion-based method, PerCo~\cite{careil2023towards}, leverages the strong generative priors of diffusion models, achieving higher performance in detection tasks compared to image coding for machine methods~\cite{chen2023transtic,li2024image}. However, due to its general design rather than a specific focus on machine vision, PerCo shows limitations in segmentation and classification. In contrast, our method employs semantic disentanglement to encode information according to the semantic requirements of each machine vision task. This targeted approach reduces the bits required for encoding information irrelevant to downstream tasks, resulting in improved performance across detection, segmentation, and classification tasks compared with other methods. Specifically, setting VTM-12.1 as the anchor, we can achieve BD-rates~\cite{Bjntegaard2001CalculationOA} of -80.41\%, -80.32\%, and -77.63\% on these three tasks, respectively.

\subsubsection{Human Perception}
% \vspace{-2mm}
While achieving strong performance in machine vision tasks, our method also maintains high-quality human perception. As shown in the two subplots in the upper-left corner of Fig.~\ref{fig:human}, our method, after disentanglement for the detection task, can still reconstruct high-quality images (low FID and DISTS, i.e., -45.615 BD-FID and -0.123 BD-DISTS~\cite{xu2023conditional} by setting VTM-12.1 as the anchor) on the decoding side, outperforming both the generative method PerCo~\cite{careil2023towards} and ILLM~\cite{muckley2023improving}. 

Moreover, by transmitting the full bit-stream, we can further enhance human perception quality. As demonstrated in Fig.~\ref{fig:human}, our method achieves superior human perception performance on the Kodak and CLIC datasets, outperforming other methods in FID, DISTS, and KID. Although the LPIPS of our method is slightly higher than that of ILLM due to the stochastic characteristic of diffusion, it remains superior to the counterpart diffusion-based method PerCo.

\subsection{Ablation Study}
\begin{figure}[t]
    \centering
    \includegraphics[width=1\linewidth]{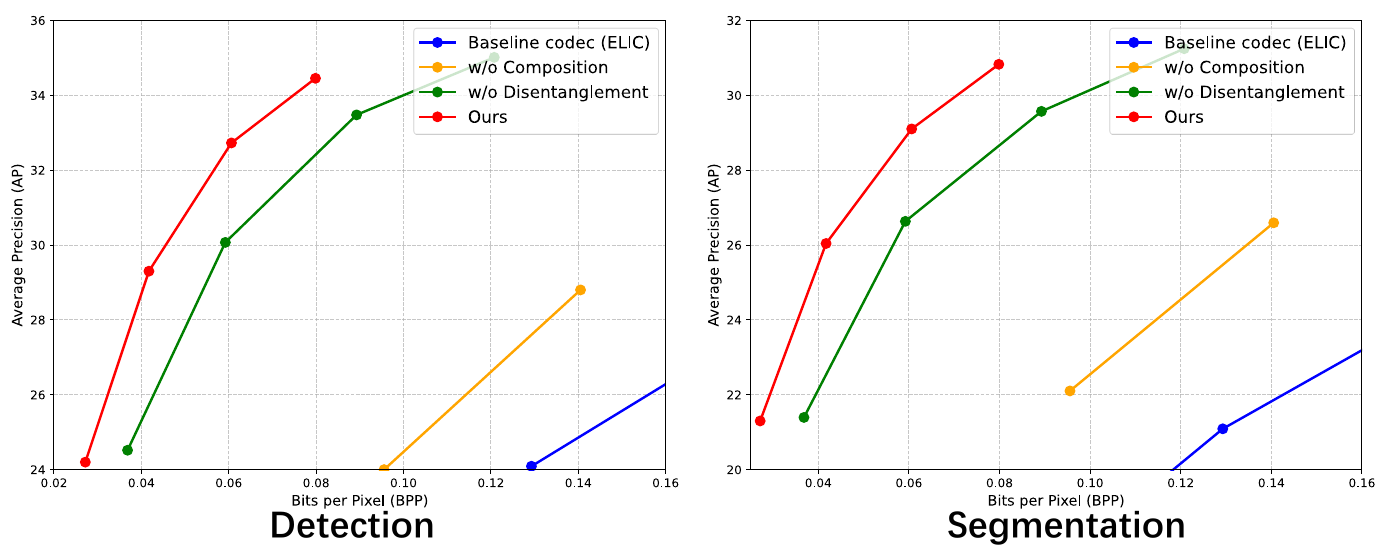}
    \caption{Ablation studies on COCO dataset. ``w/o composition" means that on the decoder side, the generative model is not used and the reconstructed image is directly output through \( g_s \). ``w/o disentanglement" indicates that the latent code is not disentangled according to the task, and the full bit-stream is transmitted.}
    \label{fig:ablation}
\end{figure}

In Fig.~\ref{fig:ablation}, we analyze the gains achieved through semantic disentanglement and generative enhancement. It can be observed that directly applying disentanglement without generative composition provides some performance improvements over the base codec. However, as the base codec is optimized for MSE rather than designed specifically for machine vision, and conducting disentanglement alone results in the loss of some contextual information, these factors together lead to performance bottlenecks in machine vision tasks. On the other hand, using generative composition alone yields substantial gains, yet since it is optimized for human eyes, and retains a considerable amount of redundant information for machine tasks. Our method leverages both disentanglement and composition processes, achieving further improvements in machine vision task performance, and also saving the bit costs.

\subsubsection{Diffusion Steps}
\label{sec:step}
\begin{figure}
    \centering
    \includegraphics[width=1\linewidth]{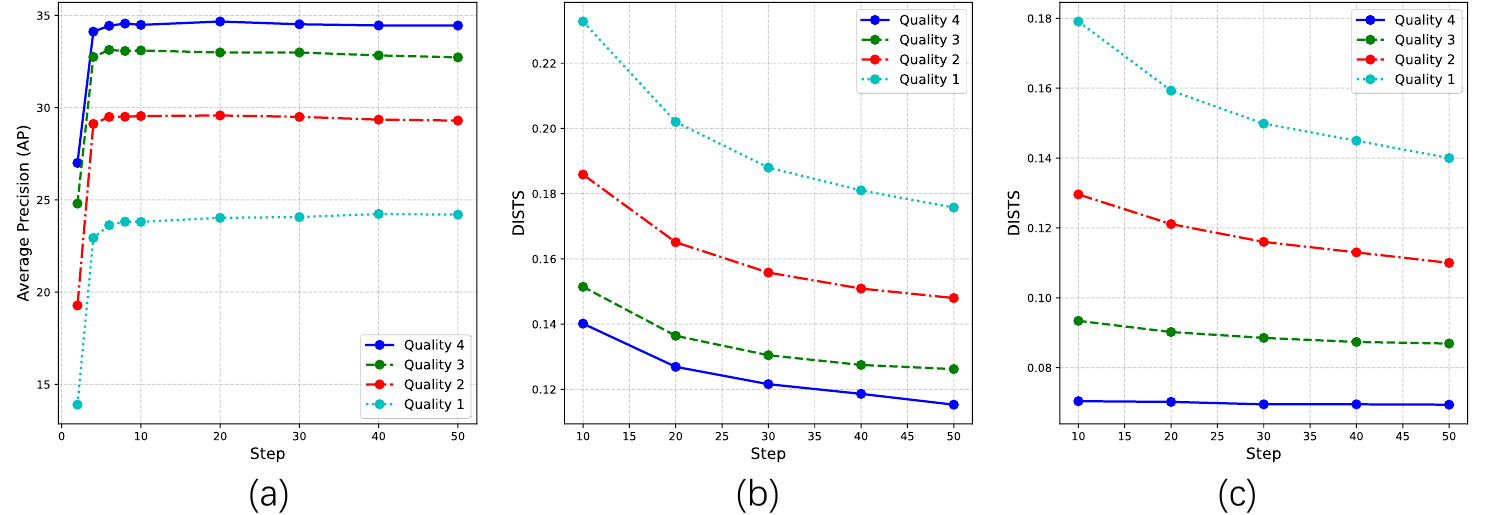}
    \caption{Ablation studies of diffusion steps on the (a) object detection task, (b) image reconstruction (Using detection labels codebook), and (c) image reconstruction (Full). Quality levels 1–4 represent progressively higher bitrates. UniCodec achieves significant performance improvements with only 2 sampling steps on machine tasks.}
    \label{fig:step}
\end{figure}

As shown in Fig.~\ref{fig:step}, we illustrate the impact of diffusion sampling steps on UniCodec's performance for both machine vision and human perception. As shown in Fig.~\ref{fig:step}(a), UniCodec achieves significant performance improvements with only 2 sampling steps on machine tasks. Stable machine vision performance is attained with 8 sampling steps. This ensures that UniCodec can handle machine vision tasks without incurring excessive decoding time overhead due to an unnecessarily high number of sampling steps.
For human perception, when only transmitting task-related bitstreams, as shown in Fig.~\ref{fig:step}(b), more steps are required to compensate for missing detail information. Conversely, as shown in Fig.~\ref{fig:step}(c), when transmitting the full bitstream, less generated information is needed. In high-bitrate scenarios, optimal performance is achieved within 10 steps.

\subsubsection{Generated Labels by LLMs}
\label{sec21}
\begin{figure}
    \centering
    \includegraphics[width=1\linewidth]{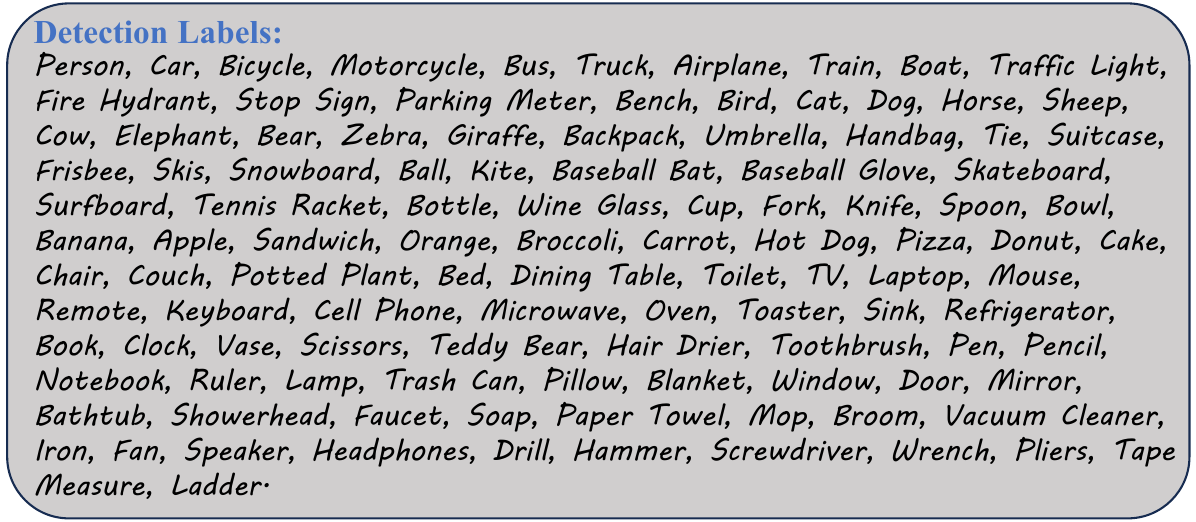}
    \caption{The generated labels of the object detection task.}
    \label{fig:objlabel}
\end{figure}

\begin{figure}
    \centering
    \includegraphics[width=1\linewidth]{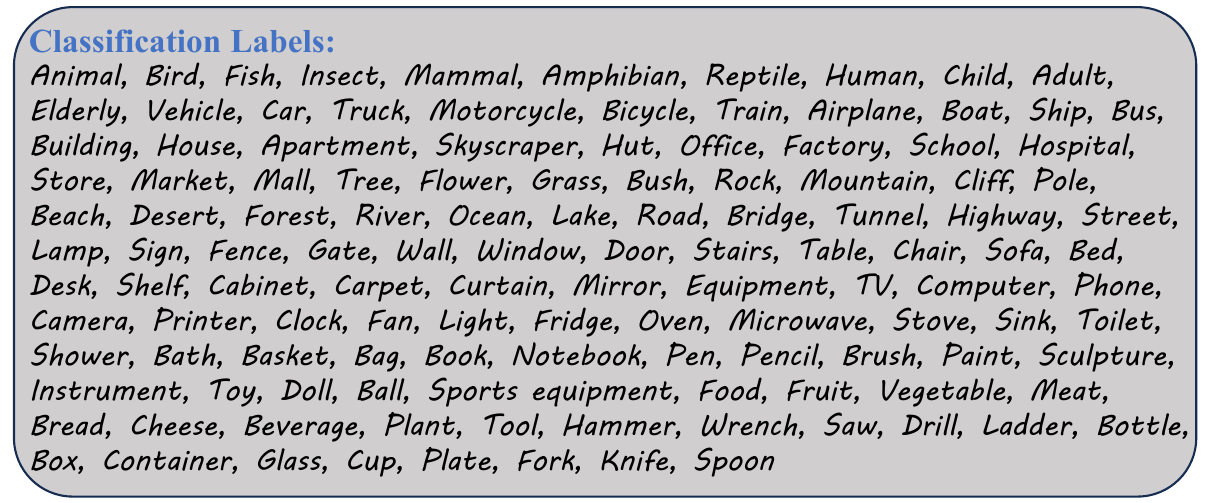}
    \caption{The generated labels of the image classification task. Compared to object detection, it includes more coarse-grained labels, such as 'animal' and 'vehicle', to ensure comprehensive coverage.}
    \label{fig:clslabel}
\end{figure}

\begin{figure}
    \centering
    \includegraphics[width=1\linewidth]{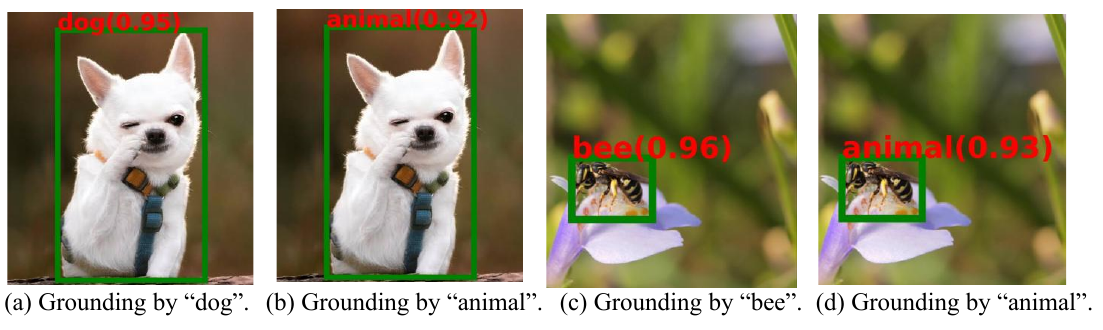}
    \caption{The visualization results by using different granularity label. The use of coarse-grained labels allows for broader object coverage.}
    \label{fig:dog}
\end{figure}

We present the labels generated for detection and classification tasks, as shown in Fig.~\ref{fig:objlabel} and~\ref{fig:clslabel}, respectively. It can be observed that these labels encompass a wide range of categories. Notably, for the classification task, some labels are generated with relatively coarse granularity, such as \textit{animal} and \textit{insect}, to ensure comprehensive coverage of the label set. For example, as shown in Fig.~\ref{fig:dog}, even without the specific labels "dog" and "bee," we can still detect the corresponding objects by using the general "animal" label, thereby reducing the likelihood of missed objects.

\begin{figure}
    \centering
    \includegraphics[width=1\linewidth]{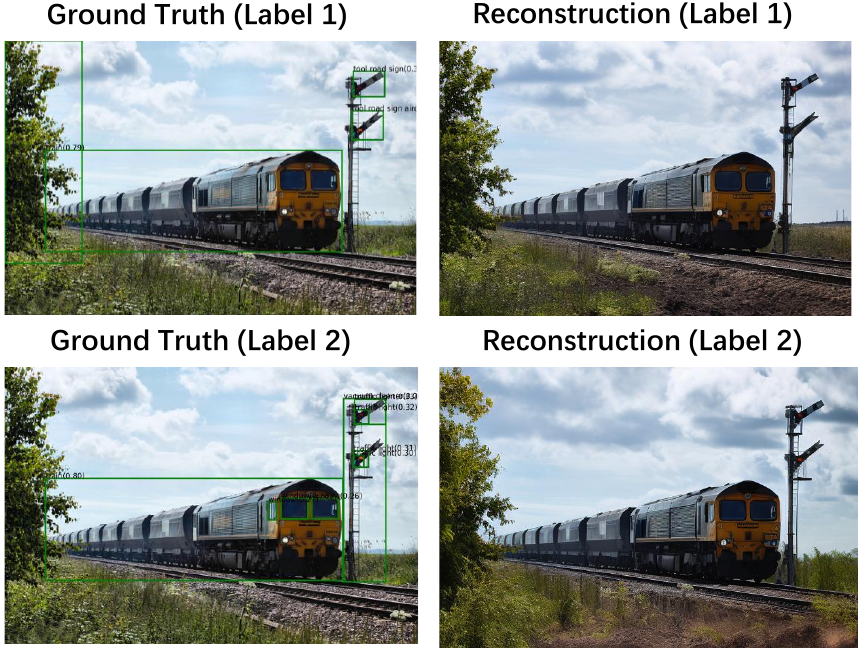}
    \caption{The visualization results by using different labels. Even if the 'tree' is not detected in label2, the generative model and global information can complete the missed object.}
    \label{fig:random_vis}
\end{figure}

It is important to note that, although the generated labels cover a broad spectrum of categories, some rare categories may remain uncovered. Nevertheless, our diffusion-based codec can mitigate this limitation by generating priors and utilizing side information to compensate for the missing details. This approach ensures that the final performance does not experience a significant drop despite these omissions. 
Fig.~\ref{fig:random_vis} and Fig.~\ref{fig:labelnum} provide examples to illustrate this situation. It can be observed that, despite using two different label sets, Label1 and Label2, our model is able to reconstruct the overall structure of the original image. Objects not included within the labeled bounding boxes, while differing in detailed textures from the original image, still retain sufficient semantic information.

\subsubsection{The Numbers of Label Generation}
\begin{figure}
    \centering
    \includegraphics[width=1\linewidth]{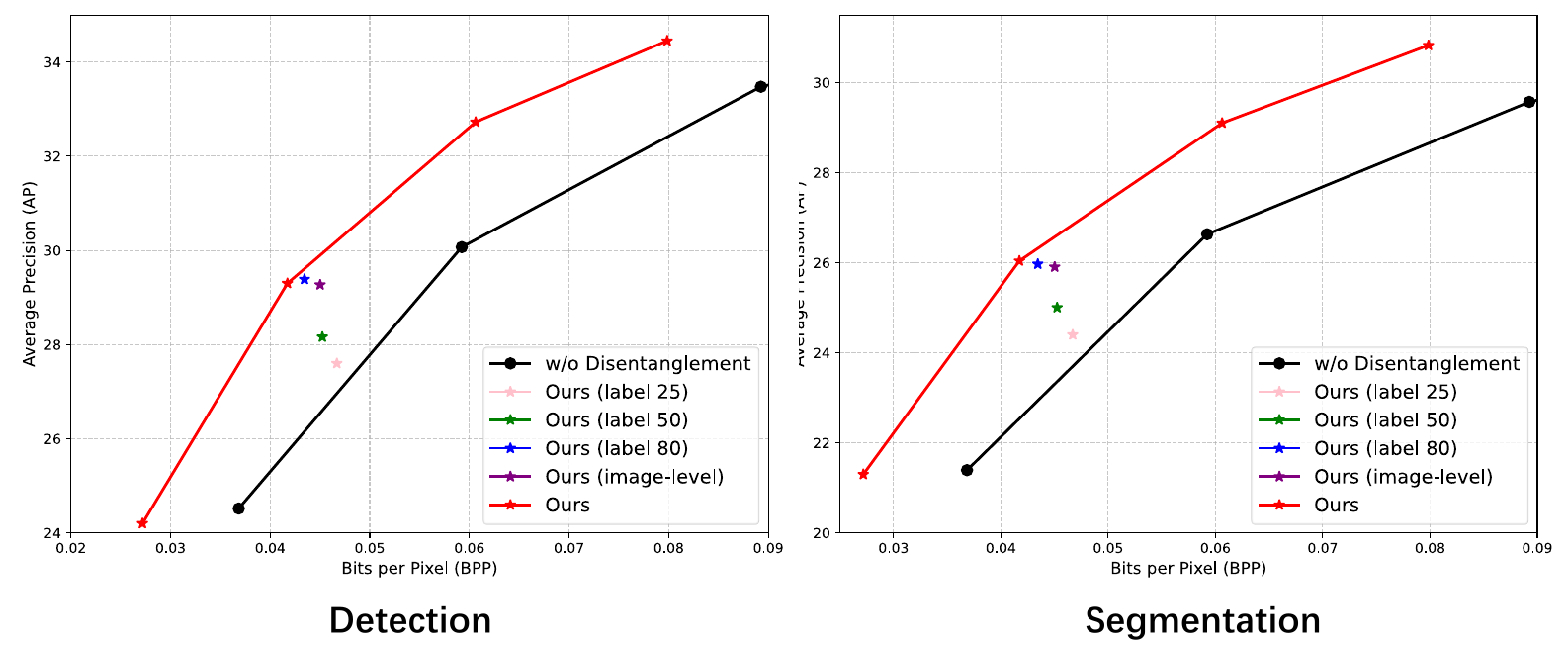}
    \caption{The performance comparison by using different numbers of labels. An increase in the number of labels leads to a performance improvement. Moreover, our method outperforms the use of MLLMs for image-level inference.}
    \label{fig:labelnum}
\end{figure}

As shown in Fig.~\ref{fig:labelnum}, we test the impact of generating different numbers of labels on the performance of machine vision tasks. It can be observed that as the number of labels increases, performance gradually improves compared with "w/o disentanglement". Considering the performance upper bound of LLM models~\cite{chatgpt_2024} and visual grounding models~\cite{liu2023grounding}, we set the number of labels to exceed 100 when designing the prompt.

We also compared our task-level LLM reasoning with image-level MLLM reasoning. For image-level MLLM reasoning, we attempted to use a locally deployed open-source multimodal large model (InternVL2)~\cite{chen2023internvl} to generate image-level labels for each image. However, due to performance gaps between open-source and closed-source large models, issues such as hallucinations and repetitive labels arose when generating labels. This resulted in the performance of image-level labels being inferior to that of task-level labels, as shown in Fig.~\ref{fig:labelnum}.

\subsubsection{Discussion on the Randomness of Label Generation}
\begin{figure}
    \centering
    \includegraphics[width=0.8\linewidth]{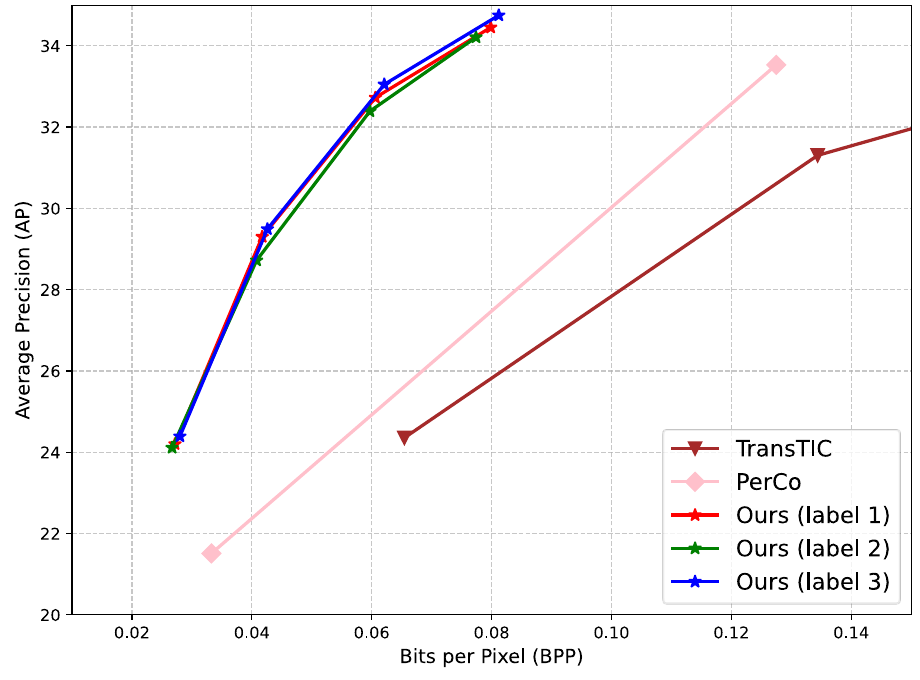}
    \caption{The results by using different label sets. Even though there is some randomness in the label generation, its impact on the final performance is minimal.}
    \label{fig:random}
\end{figure}

Due to the inherent randomness of LLMs~\cite{yang2023dawn,liu2024visual,you2023ferret,chatgpt_2024}, we conducted three separate trials to generate the same numbers of labels for the detection task to check the stability of our method. As shown in Fig.~\ref{fig:random}, the three sets of labels achieve similar performance, demonstrating the stability of our method. As discussed in Sec.~\ref{sec21}. This robustness is attributed to our semantic compositional generation module, which ensures that even if the labels exhibit slight variations, it can leverage its generative capability to produce high-quality images.

\begingroup
\subsubsection{Necessity of LLM-Grounding vs.\ Random Masking}
\label{sec:random_mask_discuss}
A natural question arises from our two-stage training design: since Stage~II uses random masks (zeroing 0--50\% of $\hat{\boldsymbol{y}}$) to train the decoder for compositional generation, why is LLM-Grounding still necessary at inference time? We clarify that these two mechanisms serve fundamentally different and complementary purposes.

\textbf{Training-time random masking} functions as a task-agnostic data augmentation. It exposes the decoder to diverse partial-input configurations, enabling it to learn a general compositional generation capability that is robust to any spatial pattern of missing information. This universality is essential: if Stage~II were trained with only LLM-Grounding-based masks, the decoder would overfit to the spatial patterns of specific grounded regions and lose the flexibility needed for different tasks and codebook configurations.

\textbf{Inference-time LLM-Grounding} provides the task-aware \textit{semantic selection} that decides \textit{which} latent regions to transmit. Given a downstream task (e.g., object detection), the LLM-generated codebook and grounding model identify the task-critical regions (e.g., vehicles, pedestrians) and allocate bits to preserve their latent representations. Without this guidance, a random mask would arbitrarily discard task-critical regions, degrading downstream performance while wasting bits on irrelevant content.

To validate this empirically, we evaluate three masking strategies at inference time on COCO object detection (Faster R-CNN, R50-FPN), as shown in Table~\ref{tab:mask_ablation}. All settings use the same trained model.

\begin{table}[t]
\centering
\small
\caption{Ablation: masking strategy at inference on COCO detection.}
\label{tab:mask_ablation}
\begin{tabular}{lcc}
\toprule
\textbf{Inference Mask Strategy} & \textbf{bpp} & \textbf{AP} \\
\midrule
w/o disentanglement (full $\hat{\boldsymbol{y}}$ transmitted) & 0.059 & 30.1 \\
Random mask & 0.053 & 19.8 \\
LLM-Grounding (our full method) & 0.060 & 32.7 \\
\bottomrule
\end{tabular}
\end{table}

The results show that random masking at inference drops AP by approximately 12.9 points compared with LLM-Grounding at the same bitrate. Moreover, even transmitting the full bitstream (w/o disentanglement) at a higher cost does not match LLM-Grounding, because the latter concentrates bits on task-relevant content. In summary, training-time random masking makes the decoder \textit{robust to any partial input}, while inference-time LLM-Grounding makes the \textit{partial input intelligent and task-aware}. Both are essential components of our framework.
\endgroup
\begin{figure}
    \centering
    \includegraphics[width=1\linewidth]{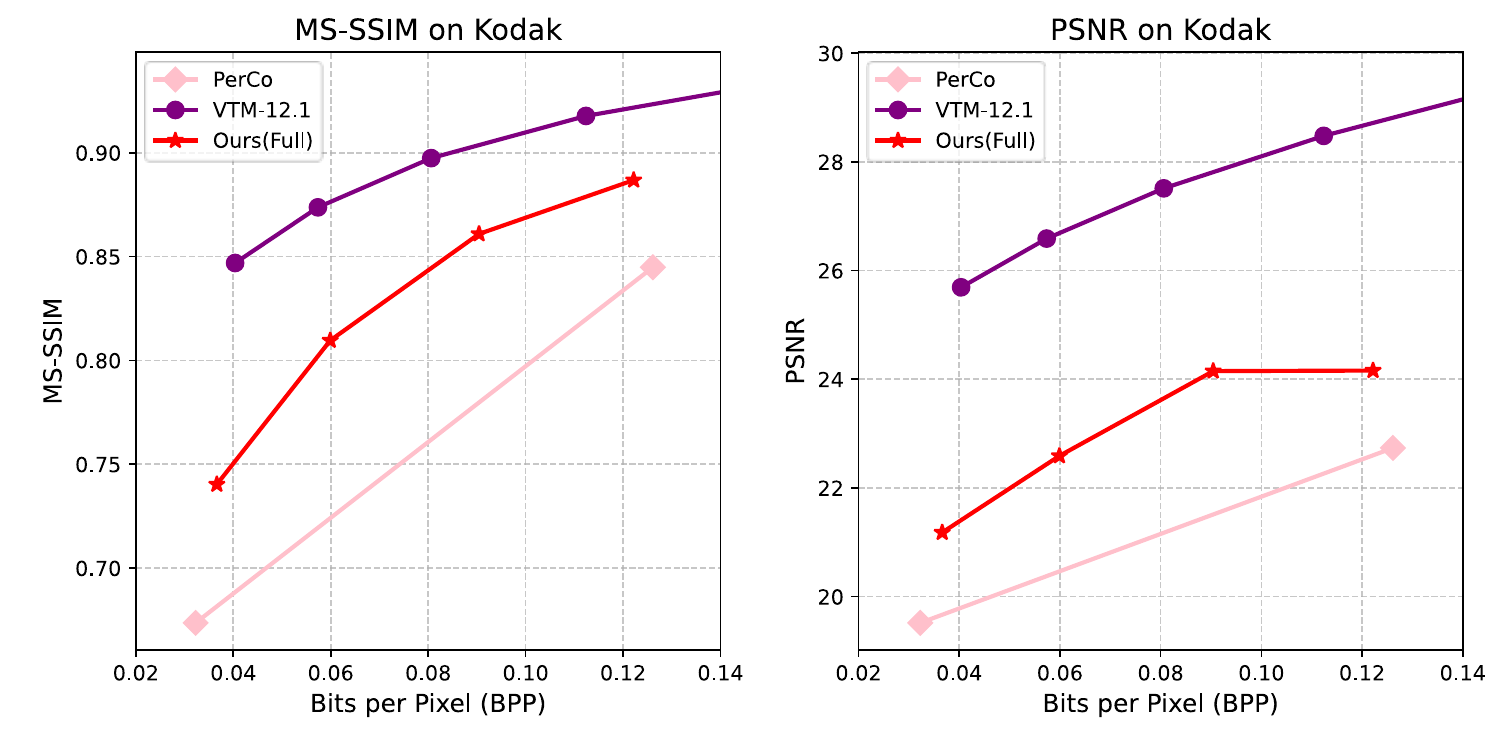}
    \caption{The comparison of fidelity metric. As a generative method, our approach scores lower on fidelity metrics than fidelity-based methods, but it outperforms the generative method PerCo~\cite{careil2023towards}.}
    \label{fig:fidelity}
\end{figure}

\subsection{Task Transferability}
\begin{figure}
    \centering
    \includegraphics[width=1\linewidth]{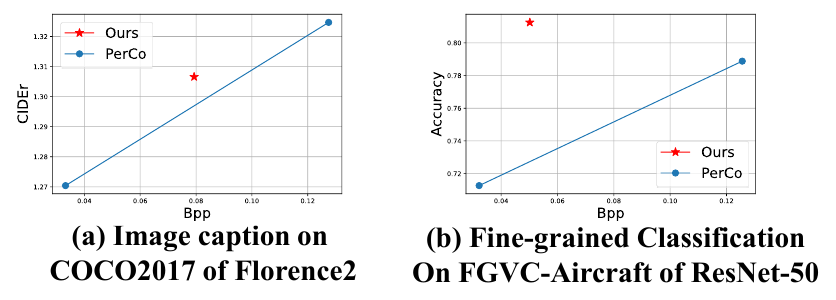}
    \caption{Performance comparison of the fine-grained classification and caption task. Our method is effective across various tasks, which demonstrates its task transferability.}
    \label{fig:transferability}
\end{figure}

To demonstrate our method's transferability, we further evaluated it on fine-grained aircraft classification and image captioning, where it also outperforms \revise{PerCo}~\cite{careil2023towards}, as illustrated in Fig.~\ref{fig:transferability}.
\revise{These transfer experiments already indicate that the encoder-side semantics do not merely overfit to the three main benchmarks, but instead remain useful when the downstream task changes.}

\subsection{Inference Time}
In Table~\ref{tab:time}, we present the inference time of our method. Using 2 steps, our approach achieves inference speeds comparable to non-generative methods~\cite{li2023frequency,vvc} while delivering superior performance on machine vision tasks, as discussed in Section~\ref{sec:step}. Furthermore, under the same number of steps, our method demonstrates faster inference times compared to other generative approaches~\cite{careil2023towards,lei2023text+}.

% Please add the following required packages to your document preamble:
% \usepackage[table,xcdraw]{xcolor}
% Beamer presentation requires \usepackage{colortbl} instead of \usepackage[table,xcdraw]{xcolor}
% Please add the following required packages to your document preamble:
% \usepackage{multirow}
% \usepackage[table,xcdraw]{xcolor}
% Beamer presentation requires \usepackage{colortbl} instead of \usepackage[table,xcdraw]{xcolor}
\begin{table}[t]
\setlength{\tabcolsep}{1mm}
\footnotesize
\caption{Inference time comparison. Using 2 steps, our approach achieves inference speeds comparable to non-generative methods.}
\begin{tabular}{l|l|l|l|l}
\hline
\textbf{Method}                 & \textbf{Steps} & \textbf{Encoding Time}                & \textbf{Decoding Time}                & \textbf{Platform}                    \\ \hline
\textbf{VTM~\cite{vvc}}                    & -              & {\color[HTML]{000000} 33.447±1.32}    & {\color[HTML]{000000} 0.131±0.02}   & {\color[HTML]{000000} AMD 7V13} \\ \hline
\textbf{Frequency~\cite{li2023frequency}}        & -              & {\color[HTML]{000000} 0.128 ± 0.008}  & {\color[HTML]{000000} 0.136 ± 0.006}  & {\color[HTML]{000000} A100}          \\ \hline
\textbf{PICS~\cite{lei2023text+}}                   & 25             & {\color[HTML]{000000} 62.045 ± 0.516} & {\color[HTML]{000000} 12.028 ± 0.413} & {\color[HTML]{000000} RTX4090}       \\ \hline
\textbf{PerCo~\cite{careil2023towards}}                  & 20             & {\color[HTML]{000000} 0.080 ± 0.018}  & {\color[HTML]{000000} 2.551 ± 0.018}  & {\color[HTML]{000000} A100}          \\ \hline
                                & 2              & {\color[HTML]{000000} 0.163 ± 0.011}    & {\color[HTML]{000000} 0.231 ± 0.007}    & {\color[HTML]{000000} A100}          \\ \cline{2-5} 
                                & 8              & {\color[HTML]{000000} 0.163 ± 0.011}    & {\color[HTML]{000000} 0.721 ± 0.007}    & {\color[HTML]{000000} A100}          \\ 
                                \cline{2-5} 
                                & 20              & {\color[HTML]{000000} 0.163 ± 0.011}    & {\color[HTML]{000000} 1.762 ± 0.008}    & {\color[HTML]{000000} A100}          \\ \cline{2-5} 
\multirow{-3}{*}{\textbf{Ours}} & 50             & {\color[HTML]{000000} 0.163 ± 0.011}    & {\color[HTML]{000000} 3.960 ± 0.009}     & {\color[HTML]{000000} A100}          \\ \hline
\end{tabular}
\label{tab:time}
\end{table}

\subsection{Visualization}
\begin{figure*}[!ht]
    \centering
    \includegraphics[width=1\linewidth]{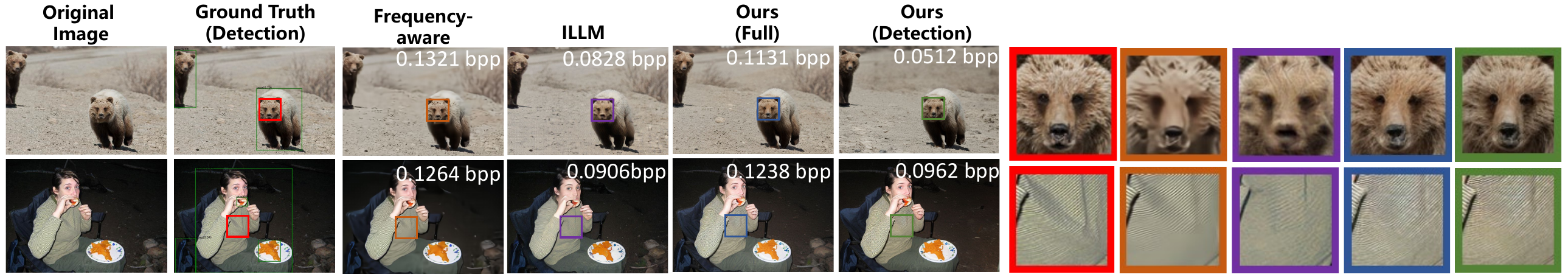}
    \caption{The visualization results. UniCodec preserves information in task-related regions for machine vision tasks while effectively meeting human visual perception requirements. The scheme of ``Ours (Detection)'' means the reconstructed image of our method by transmitting only the partial task-related bit-stream, while ``Ours (Full)'' means the full bit-stream is transmitted. ``Ground Truth (Detection)'' means the results of grounding model, which indicates the task-related regions for the object detection task. }
    \label{fig:visual}
\end{figure*}

In Fig.~\ref{fig:visual}, we present some visualized results. The figure shows that the fidelity-based method, Frequency-aware~\cite{li2023frequency}, results in blurring, particularly in complex textures, such as the striped clothing in the second row; it struggles to reconstruct details accurately. In contrast, our approach leverages the priors from diffusion, enabling more detailed reconstruction compared with the generative method ILLM~\cite{muckley2023improving}. For example, the bear's face in the first row shows significant distortion in both ILLM and fidelity-based methods, while our method preserves more details.

Furthermore, we compared the reconstruction quality of transmitting only the task-related region bit-streams, Ours (Detection), versus the full bit-streams, Ours (Full). 
For task-related regions, both approaches achieve similar results indicating task-related information is well preserved. Additionally, by transmitting only the task-related region bit-stream, our method, with the help of diffusion priors, can reconstruct approximate background information, satisfying human perception, while reducing substantial bits.

\subsection{Progressive Coding}
\begin{figure*}
    \centering
    \includegraphics[width=1\linewidth]{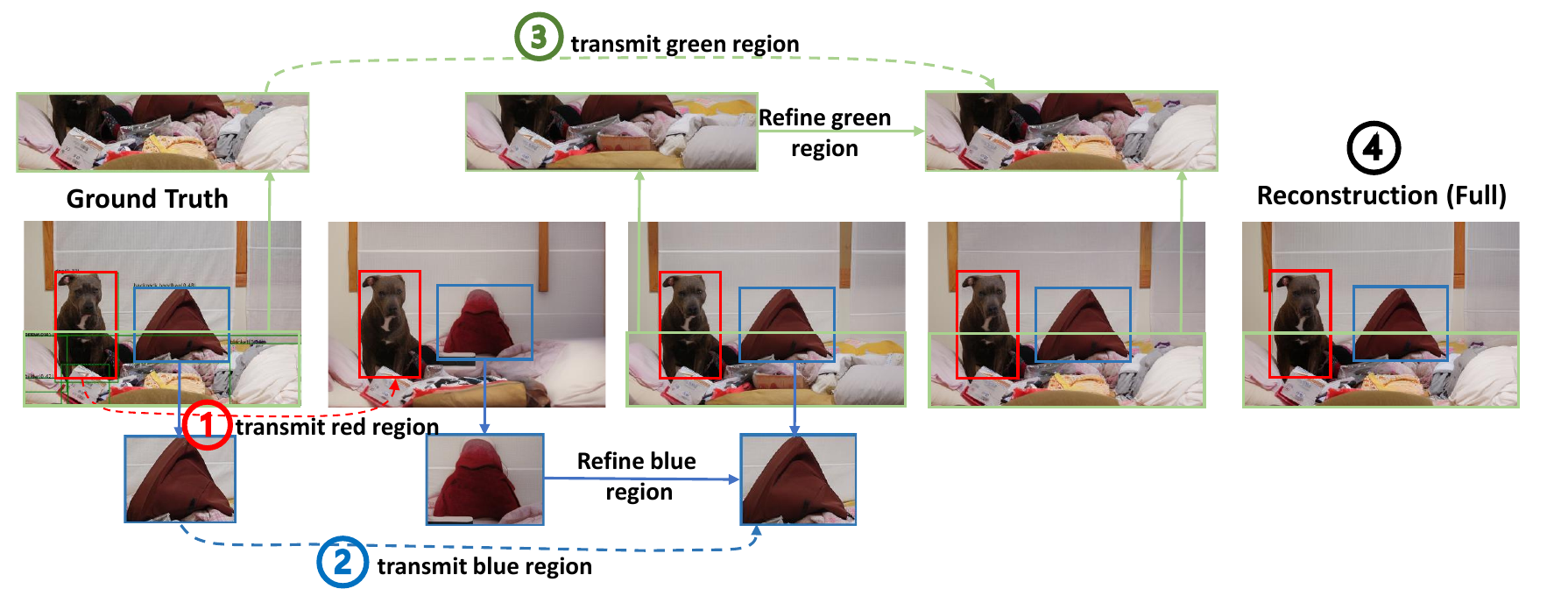}
    \caption{The visualization results of progressive coding. We transmit the \textcolor{red}{red}, \textcolor{blue}{blue}, \textcolor{green}{green} regions, and the remaining regions in sequence for reconstruction. Our method can reconstruct a complete, high-quality image by transmitting only the red region, and can further refine it by progressively transmitting the other regions.}
    \vspace{-3mm}
    \label{fig:progressive}
\end{figure*}

In Fig.~\ref{fig:progressive}, we illustrate the step-by-step transmission of different regions and the corresponding reconstruction results of UniCodec. It can be observed that the transmitted regions maintain a high level of consistency with the original image. Furthermore, even with the transmission of only a small portion of regions, UniCodec is still capable of reconstructing a complete image that aligns well with human visual perception.
\revise{The semantic regions used for progressive transmission are non-overlapping by construction, so the transmitted latent tiles are not redundantly counted across steps.}

\section{Limitation}
Since our method is designed for both human perception and machine vision tasks, it tends to preserve semantic information but not fidelity. As a result, it may perform a little worse on certain fidelity metrics, i.e., PSNR and MS-SSIM, compared with the compression standard VVC in Fig.~\ref{fig:fidelity}. However, compared with other diffusion-based methods, Perco~\cite{careil2023towards}, the reconstructed images from our approach demonstrate a certain advantage in fidelity.

\section{Discussion}
\subsection{The role of LLM and Grounding Model}
A key aspect of our methodology is the two-stage semantic analysis, where the Large Language Model (LLM) plays a foundational role. By generating a comprehensive, reusable label codebook for each specific application, the LLM endows our framework with task versatility and efficiency, eliminating the need for costly retraining and the high latency associated with per-image reasoning.

Following this high-level analysis, the grounding model at the encoder performs image-specific, task-aware disentanglement. This process informs the codec by identifying semantically salient regions relevant to the given task, enabling a more efficient allocation of bits. It is essential to note that the final task performance is evaluated on the reconstructed image produced by the decoder. The disentangled information from the encoder serves exclusively as input to guide the compression and subsequent generative reconstruction; it is not passed to the final evaluation network. The efficacy and generality of this semantically-guided compression are demonstrated by UniCodec's strong performance across a diverse set of downstream applications, including object detection, segmentation, classification, fine-grained classification, and image caption. The framework's ability to excel in these varied tasks, each with distinct semantic requirements, underscores that our approach provides a genuine and broadly applicable enhancement to the compression process. This versatility confirms that the method improves the overall semantic fidelity of the reconstructed image, making it highly effective for a wide range of machine vision uses.
\revise{We further clarify the codec-side novelty beyond the pretrained diffusion decoder. The first component is task-conditioned latent disentanglement driven by reusable codebooks and open-vocabulary grounding. The second is a structured bitstream that separates $\tilde{\boldsymbol{y}}_i$ and $\hat{\boldsymbol{z}}$, enabling independent control of task-relevant local semantics and coarse global context. The third is the latent-space training interface that aligns the codec decoder with the pretrained diffusion/VAE space under partial transmission. Even when the same pretrained diffusion prior is kept fixed, the existing ablation in Fig.~\ref{fig:ablation} shows that removing disentanglement or composition degrades controllable rate allocation and downstream performance.}

\section{Conclusion}
We introduce a new paradigm, addressing the different requirements of human visual perception and machine vision tasks. We use LLM and grounding models to analyze the semantics of tasks and images, assisting with semantic disentanglement encoding. By leveraging semantic disentanglement, UniCodec enables partially encoding that selectively transmits task-related information, significantly reducing bits cost. The following semantic composition decoding approach utilizes the diffusion model to further enhance reconstruction, composing essential object details with prior side knowledge to generate high-quality images and achieve both satisfactory human perception and diverse machine vision task performance. Experimental results underscore the robustness and versatility of our approach, demonstrating superior performance in both machine vision and human perception evaluations. 

\section*{Acknowledgments}
This work was supported by Grants of NSFC 62302246, ZJNSFC LQ23F010008, Ningbo 2023Z237 \& 2024Z284 \& 2024Z289 \& 2023CX050011 \& 2025Z038 \& 2025Z059, and supported by High Performance Computing Center at Eastern Instituteof Technology and Ningbo Institute of Digital Twin. \looseness=-1
% \clearpage
% \clearpage

\bibliographystyle{IEEEtran}
\bibliography{main}

\section{Biography Section}
\begin{IEEEbiography}[{\includegraphics[width=1in,height=1.25in,clip,keepaspectratio]{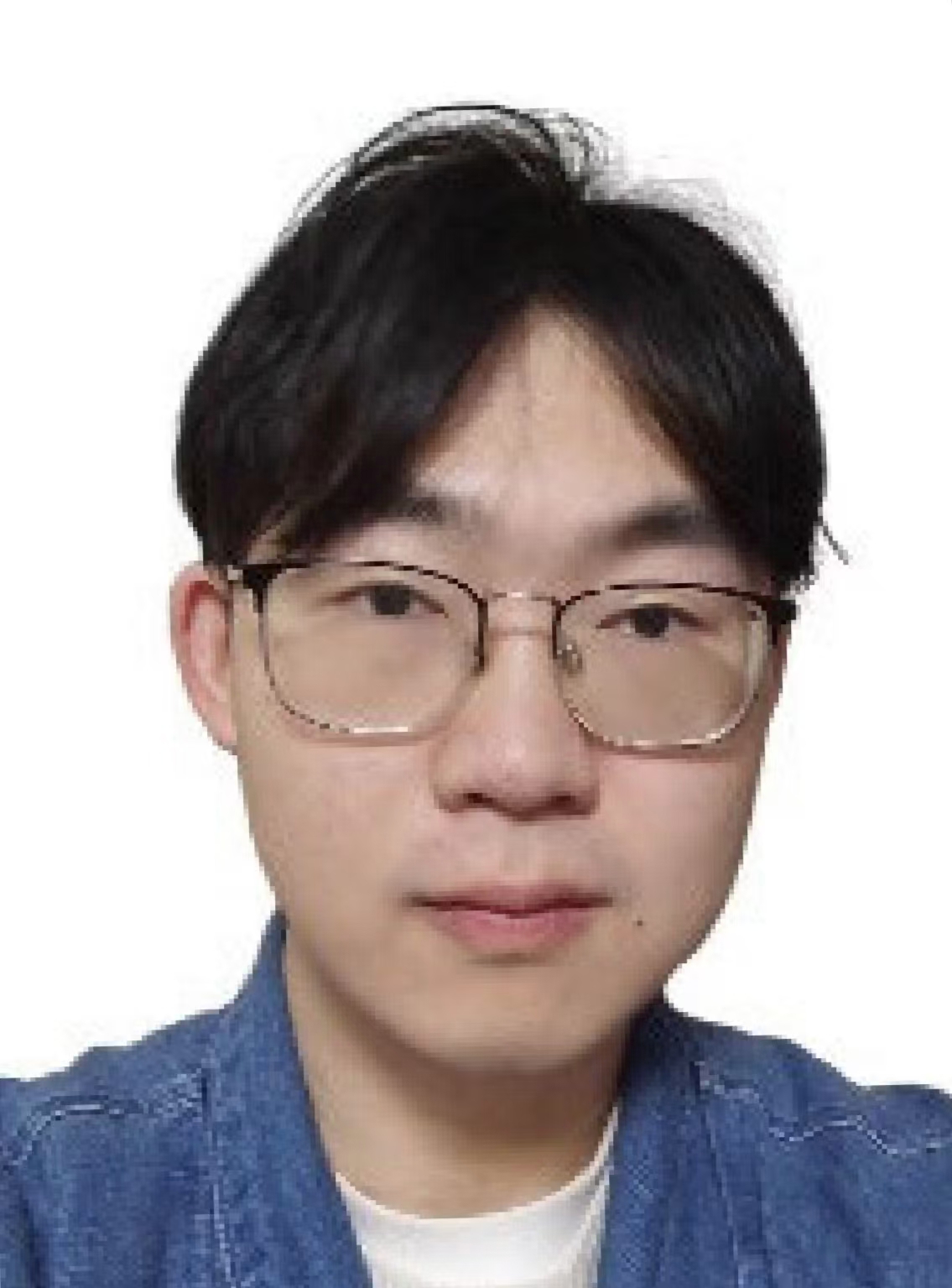}}]{Jinming Liu} received the B.S. degree from Southeast University and the M.Sc. degree from Waseda University. He is currently pursuing the Ph.D. degree with the School of Electronic Information and Electrical Engineering, Shanghai Jiao Tong University. His current research interests include image/video compression, computer vision, and deep learning.
\end{IEEEbiography}

\begin{IEEEbiography}[{\includegraphics[width=1in,height=1.25in,clip,keepaspectratio]{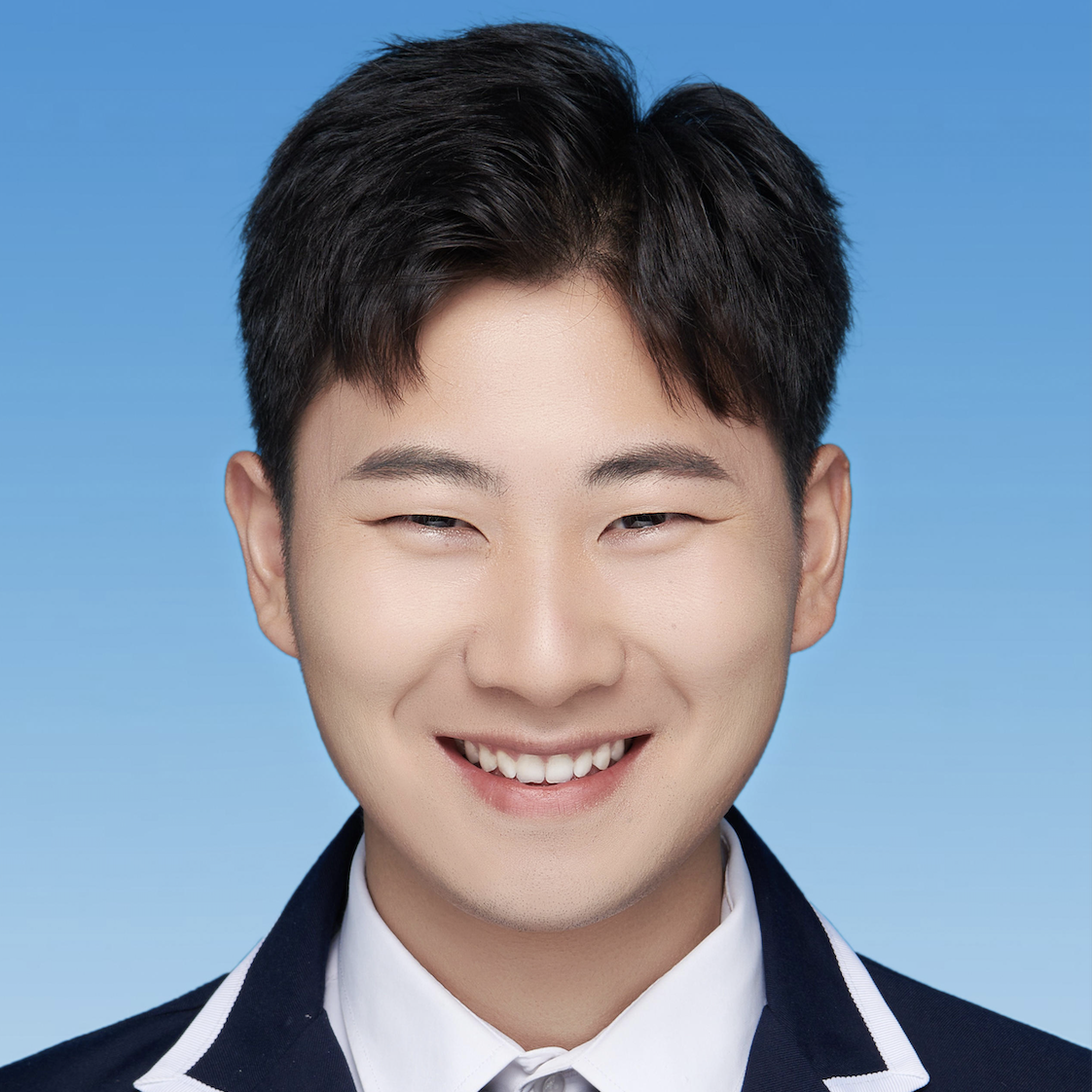}}]{Yuntao Wei}
is currently a joint Ph.D. student from the Department of Computing of The Hong Kong Polytechnic University and  Eastern Institute of Technology, Ningbo. He received his M.E. degree from the School of Artificial Intelligence and Data Science, University of Science and Technology of China. His main research interests include video coding for machine and visual coding for large multi-modal models.
\end{IEEEbiography}

\begin{IEEEbiography}[{\includegraphics[width=1in,height=1.25in,clip,keepaspectratio]{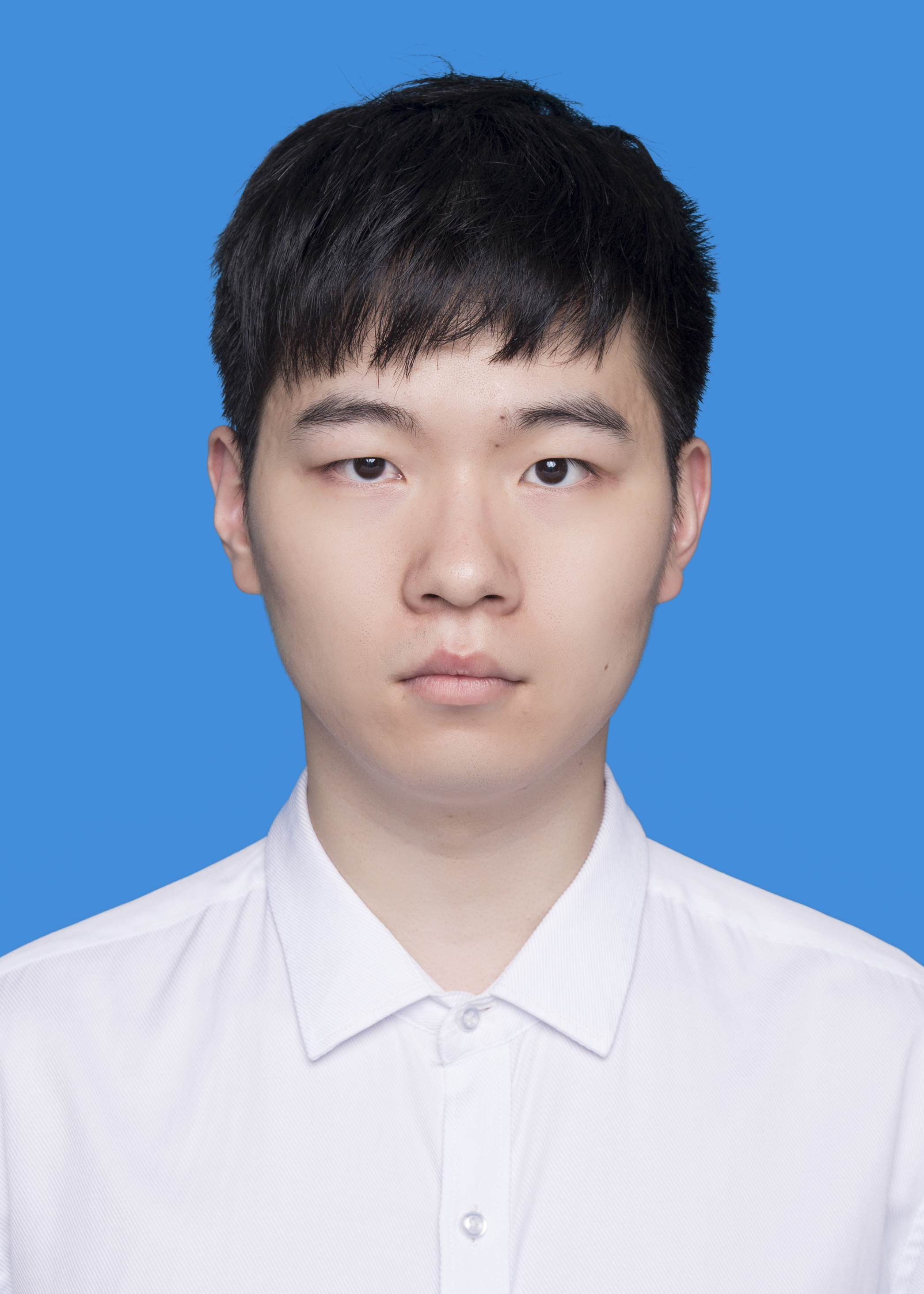}}]{Junyan Lin}
 received the B.Sc. degree in Computer Science and Technology from Zhejiang Gongshang University, Hangzhou, China, in 2022, and the M.Sc. degree in Computer Science and Technology from the School of Computer Science, Ocean University of China, Qingdao, China, in 2025. His research interests include multimodal large language models and remote sensing.
\end{IEEEbiography}

\begin{IEEEbiography}[{\includegraphics[width=1in,height=1.25in,clip,keepaspectratio]{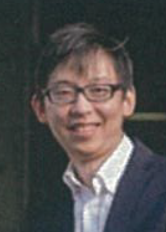}}]{Heming Sun}
received the B.E. degree in electronic engineering from Shanghai Jiao Tong University, Shanghai, China, in 2011, and received the M.E. degree from Waseda University and Shanghai Jiao Tong University, in 2012 and 2014, respectively, through a double-degree program. In 2017 he earned his Ph.D. degree from Waseda University. He was a researcher at NEC Central Research Laboratories from 2017 to 2018. He was an Assistant Professor at Waseda University from 2018 to 2023. He is now Associate Professor at Yokohama National University.
\end{IEEEbiography}

\begin{IEEEbiography}[{\includegraphics[width=1in,height=1.25in,clip,keepaspectratio]{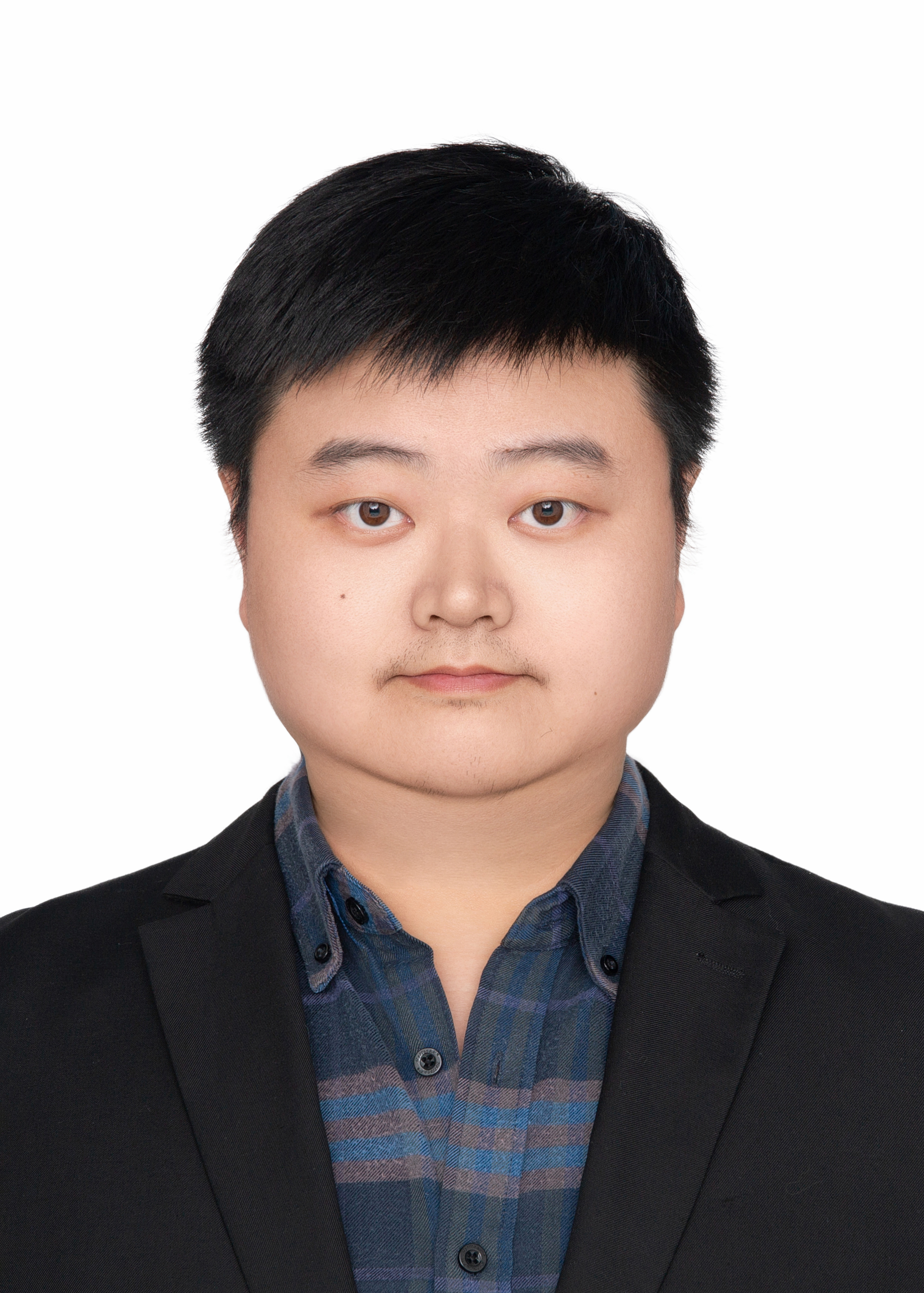}}]{Shengyang Zhao} is now an associated researcher at the Ningbo Institute of Digital Twin. He recieved his Ph.D degree in Electronic Engineering and Information Science from the University of Science and Technology of china (USTC). He is also an active member in JPEG Pleno Ad Hoc group. His research interests include immersive multimedia coding, image quality assessment, and autonomous driving.
\end{IEEEbiography}

\begin{IEEEbiography}[{\includegraphics[width=1in,height=1.25in,clip,keepaspectratio]{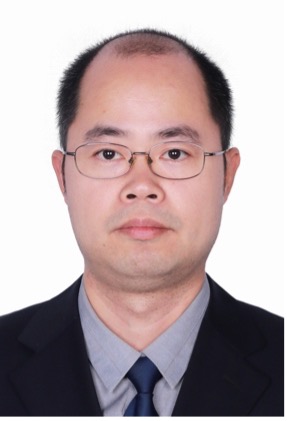}}]{Zhibo Chen}
(M’01-SM’11) received the B. Sc., and Ph.D. degree from Department of Electrical Engineering Tsinghua University in 1998 and 2003, respectively. He is now a full professor in University of Science and Technology of China. His research interests focus on investigating artificial intelligence technique for advanced visual signal generation, representation, processing and coding, as well as in other interdisciplinary research fields. He has more than 200 publications and over 100 granted patent applications. Some of his standard proposals have been adopted in MPEG/VCEG on video coding and ITU-T P.1202 on video quality assessment. 
He has served as Chair of the Technical Committee of IEEE CASS Visual Signal Processing and Communications (VSPC)
from 2021 to 2024, Senior Associate Editor of IEEE Trans. on Circuits and Systems on Video Technology, Corresponding Guest Editor of IEEE JETCAS and IEEE OJCAS. He is currently also a member of IEEE SP Image, Video, 
and Multidimensional Signal Processing (IVMSP) Technical Committee, member of the Technical Committee of IEEE 
CASS Multimedia System Applications (MSA) and a member of the Technical Committee of IEEE CASS Circuit and System Education and Outreach (CASEO).
\end{IEEEbiography}
\vfill

\begin{IEEEbiography}[{\includegraphics[width=1in,height=1.25in,clip,keepaspectratio]{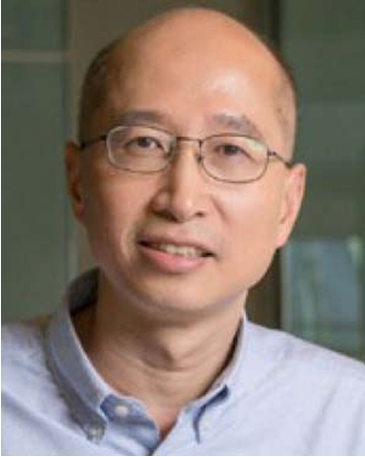}}]{Wenjun Zeng}
(Fellow, IEEE) received the B.E. degree from Tsinghua University, Beijing, China, in 1990, the M.S. degree from the University of Notre Dame, Notre Dame, IN, USA, in 1993, and the Ph.D. degree from Princeton University, Princeton, NJ, USA, in 1997. Since 2021, he has been a Chair Professor and the Vice President for Research at the Eastern Institute for Advanced Study (EIAS)/Eastern Institute of Technology (EIT), Ningbo, China. He is also the Founding Executive Director of the Ningbo Institute of Digital Twin. From 2014 to 2021, he was a Senior Principal Research Manager and a Member of the Senior Leadership Team with Microsoft Research Asia, Beijing, where he led the video analytics research empowering the Microsoft Cognitive Services, Azure Media Analytics Services, Office, and Windows Machine Learning. He was with the University of Missouri, Columbia, MO, USA, from 2003 to 2016, most recently as a Full Professor. Prior to that, he had worked for Packet Video Corp., Sharp Labs of America, Bell Labs, and Panasonic Technology. He has contributed significantly to the development of international standards (ISO MPEG, JPEG2000, and OMA). Dr. Zeng is on the Editorial Board of the International Journal of Computer Vision. He was an Associate Editor-in-Chief for IEEE Multimedia Magazine and an Associate Editor for IEEE TRANSACTIONS ON CIRCUITS AND SYSTEMS FOR VIDEO TECHNOLOGY, IEEE TRANSACTIONS ON INFORMATION FORENSICS AND SECURITY, and IEEE TRANSACTIONS ON MULTIMEDIA(TMM). He was on the Steering Committee of IEEE TRANSACTIONS ON MOBILE COMPUTING and IEEE TMM. He was the Steering Committee Chair of IEEE ICME in 2010 and 2011, and was the General Chair or TPC Chair for several IEEE conferences (e.g., ICME’2018, ICIP’2017). He was the recipient of several best paper awards.
\end{IEEEbiography}

\begin{IEEEbiography}[{\includegraphics[width=1in,height=1.25in,clip,keepaspectratio]{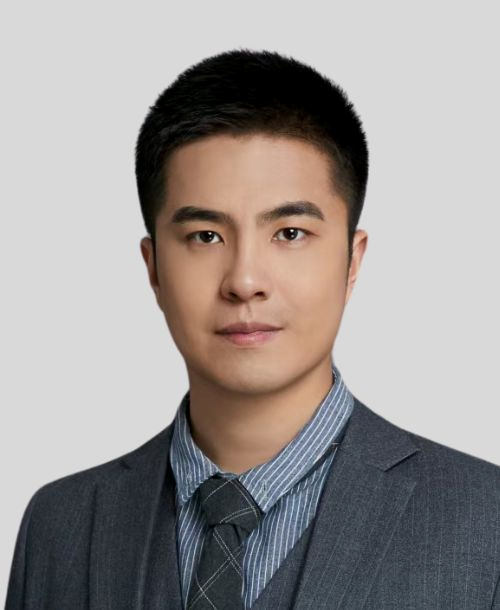}}]{Xin Jin}
(M’19) has been a tenure-track Assistant Professor with the Eastern Institute of Technology (EIT), Ningbo, China. He is also a Researcher at the Ningbo Institute of Digital Twin. He received his Ph.D. degree in Electronic Engineering and Information Science from the University of Science and Technology of China (USTC). His research interests include computer vision, intelligent media computing, and deep learning. He has over 10 granted patent applications, around 40 publications, and over 5,500 Google citations. He is an IEEE member and reviewer of IEEE Transactions on Image Processing (TIP), IEEE Transactions on Multimedia (TMM), and IEEE Transactions on Circuits and Systems for Video Technology (TCSVT). IEEE Visual Signal Processing and Communications (VSPC) TPC Member, IEEE Visual Signal Processing and Communications (VSPC) Rising Star Award 2024.
\end{IEEEbiography}
\end{document}